\documentclass[10pt,journal,compsoc]{IEEEtran}

\usepackage{graphicx}
\usepackage{booktabs}
\usepackage{subcaption}
\usepackage{epstopdf}
\usepackage{diagbox}
\usepackage{threeparttable}
\usepackage{makecell, multirow,diagbox}

\usepackage[linesnumbered,ruled,vlined]{algorithm2e}
\usepackage{algpseudocode}
\usepackage{amssymb}
\usepackage{amsmath}
\usepackage{amsthm}
\usepackage{cite}
\usepackage{hyperref}
\usepackage{bm}
\usepackage{mathtools}
\usepackage{pgfmath}
\usepackage{dsfont}
\usepackage{xcolor}
\usepackage{color}
\usepackage{lineno}
\usepackage{colortbl}
\usepackage{bbding}

\usepackage{listings}
\usepackage{soul}
\usepackage{amsmath}

\def\ourmethod{GroupNL}

\def\BibTeX{{\rm B\kern-.05em{\sc i\kern-.025em b}\kern-.08em
    T\kern-.1667em\lower.7ex\hbox{E}\kern-.125emX}}

\usepackage{flushend}

\begin{document}
%
\title{\ourmethod{}: Low-Resource and Robust CNN Design over Cloud and Device}

\author{Chuntao~Ding,
        ~Jianhang~Xie,
        ~Junna~Zhang,
        ~Salman~Raza,
        ~Shangguang~Wang, 
        ~Jiannong~Cao\\
\IEEEcompsocitemizethanks{\IEEEcompsocthanksitem Chuntao Ding is with the School of Artificial Intelligence, Beijing Normal University, Beijing, China.
E-mail: ctding@bnu.edu.cn

\IEEEcompsocthanksitem Jianhang Xie is with School of Computer Science and Technology, Beijing Jiaotong University, Beijing, China; and now is with the Department of Computer Science, City University of Hong Kong, Hong Kong, China.
E-mail: jianhang.xie@my.cityu.edu.hk.

\IEEEcompsocthanksitem Junna Zhang is with the School of Computer and Information Engineering, Henan Normal University, Xinxiang, Henan, China
\protect E-mail: jnzhang@htu.edu.cn.

\IEEEcompsocthanksitem Salman Raza is with the Department of Computer Science, National Textile University Faisalabad, Pakistan.
\protect E-mail: salmanraza@ntu.edu.pk.

\IEEEcompsocthanksitem Shangguang Wang is with the State Key Laboratory of Networking and Switching Technology, Beijing University of Posts and Telecommunications, Beijing, China.
\protect E-mail: sgwang@bupt.edu.cn.

\IEEEcompsocthanksitem Jiannong Cao is with the Department of Computing, The Hong Kong Polytechnic University, Hong Kong, China.
E-mail: {csjcao@comp.polyu.edu.hk}.

\IEEEcompsocthanksitem {Chuntao Ding and Jianhang Xie are contributed equally to this work}.
(Corresponding authors: Jianhang Xie; Junna Zhang.) 
}
}

\markboth{IEEE Transactions on Mobile Computing}%
{Shell \MakeLowercase{\textit{et al.}}: Bare Demo of IEEEtran.cls for IEEE Transactions on Magnetics Journals}

\IEEEtitleabstractindextext{%
\begin{abstract}
Deploying Convolutional Neural Network (CNN) models on ubiquitous Internet of Things (IoT) devices in a cloud-assisted manner to provide users with a variety of high-quality services has become mainstream.
Most existing studies speed up model cloud training/on-device inference by reducing the number of convolution (Conv) parameters and floating-point operations (FLOPs).
However, they usually employ two or more lightweight operations (e.g., depthwise Conv, $1\times1$ cheap Conv) to replace a Conv, which can still affect the model's speedup even with fewer parameters and FLOPs.
To this end, we propose the Grouped NonLinear transformation generation method (\ourmethod{}), leveraging data-agnostic, hyperparameters-fixed, and lightweight Nonlinear Transformation Functions (NLFs) to generate diversified feature maps on demand via grouping, thereby reducing resource consumption while improving the robustness of CNNs.
First, in a \ourmethod{} Conv layer, a small set of feature maps, i.e., seed feature maps, are generated based on the seed Conv operation.
Then, we split seed feature maps into several groups, each with a set of different NLFs, to generate the required number of diversified feature maps with tensor manipulation operators and nonlinear processing in a lightweight manner without additional Conv operations.
We further introduce a sparse \ourmethod{} Conv to speed up by reasonably designing the seed Conv groups between the number of input channels and seed feature maps.
Experiments conducted on benchmarks and on-device resource measurements demonstrate that the \ourmethod{} Conv is an impressive alternative to Conv layers in baseline models.
Specifically, on Icons-50 dataset, the accuracy of \ourmethod{}-ResNet-18 is 2.86\% higher than ResNet-18; 
on ImageNet-C dataset, the accuracy of \ourmethod{}-EfficientNet-ES achieves about 1.1\% higher than EfficientNet-ES.
In addition, we verified the efficiency of \ourmethod{}-based models in terms of cloud training and on-device inference.
\end{abstract}
}

\maketitle

\IEEEpeerreviewmaketitle

\section{Introduction} \label{ref-introduction}
\begin{figure}
\centering
\includegraphics[width=0.85\linewidth]{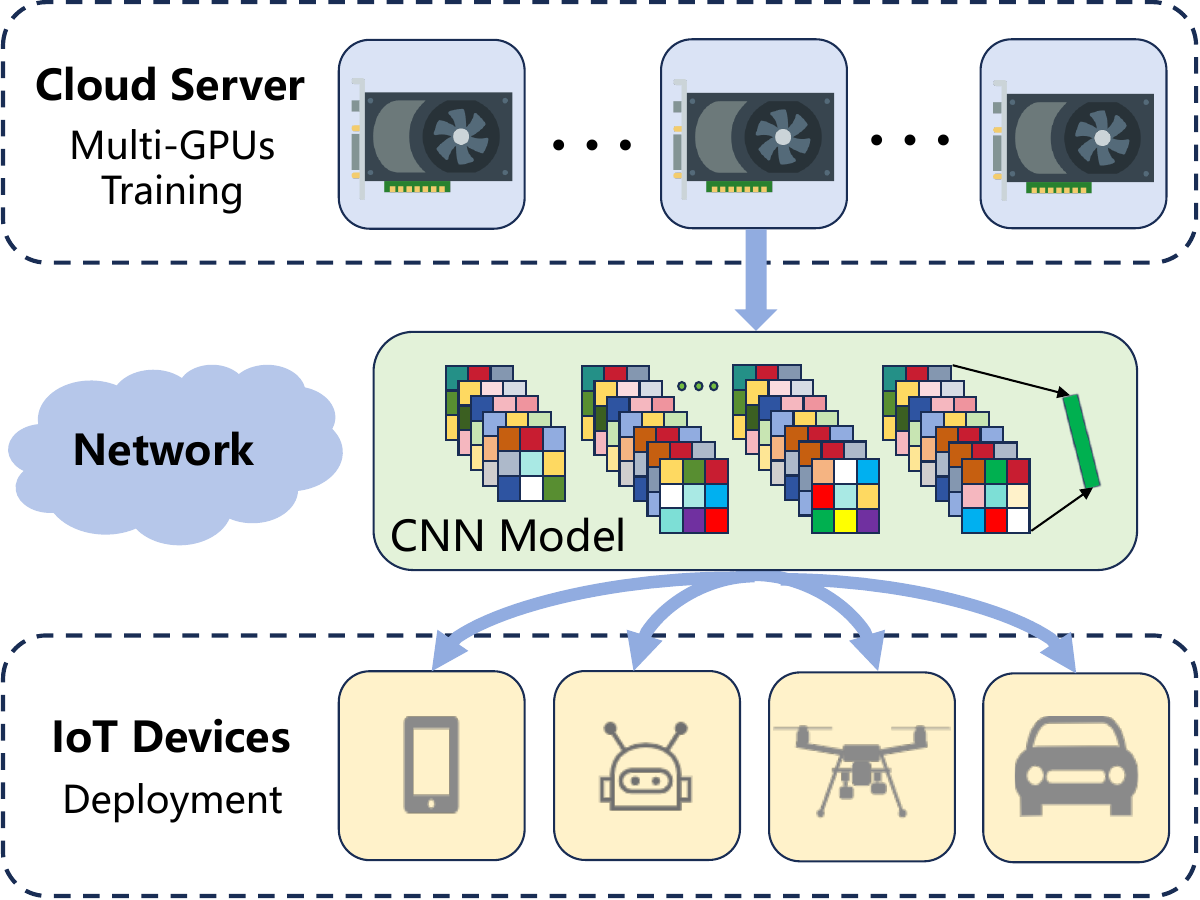}
\caption {Overview of cloud-assisted architecture.}
\label{fig:overview_problemState}
\end{figure}

\IEEEPARstart{I}{nternet} of Things (IoT) Analytics predicts that the number of connected IoT devices worldwide will increase from 16.7 billion in 2023 to 29 billion in 2027\footnote{https://iot-analytics.com/number-connected-iot-devices/}.
It is possible to provide convenient services using connected, ubiquitous IoT devices.
Due to the limited computing and storage resources of IoT devices, using the cloud server to train Convolutional Neural Netowrks (CNNs)~\cite{Karen@Very, resnet, Huang@Densely, Ding_2021_CVPR, Howard@MobileNets, mobilenetv2, Howard@Searching, efficientnet20lite, tan19efficientnet, efficientnet19edgetpu, tan21efficientnetv2} and then sending them to IoT devices for deployment and service provision (such as image recognition and monitoring services~\cite{Ding@Towards, Yu@Localized, xie2024multivision, Ping@Latency, Wang@Privacy}) has become mainstream, as shown in Fig.~\ref{fig:overview_problemState}.

Given the limited resources of IoT devices and users' demand for fast service response, many methods have been proposed to reduce the resource consumption of CNN models and speed up model training/inference by reducing the number of convolutional (Conv) parameters and floating-point operations (FLOPs) during feature map generation.
For example, GhostNet~\cite{han2020ghostnet} uses vanilla Conv and lightweight Conv (e.g., depthwise Conv~\cite{Howard@MobileNets}) to generate feature maps.
SineFM~\cite{Lu@Seed} combines vanilla Conv, nonlinear transformation functions (NLFs) with fixed hyperparameters, and $1\!\times\!1$ cheap Conv~\cite{han2020ghostnet} to generate feature maps.
These methods can reduce the number of model parameters and FLOPs, but the lightweight operations (depthwise Conv and cheap Conv) introduced affect the model's cloud training/on-device inference speed.
In addition, fixing hyperparameters of NLFs prevents model parameter updates from being completely dependent on the training data, regularizing the model to some extent and improving its robustness, while effectively enhancing the robustness of lightweight CNN models to environmental factors is a task of significant value that remains to be explored.

To this end, this paper aims to propose a CNN design method that improves model robustness by increasing feature map diversity, while reducing model complexity and avoiding the introduction of lightweight operations such as depthwise Conv and cheap Conv to achieve better training/inference speed.

First, this paper proposes the \textbf{Group}ed \textbf{N}on\textbf{L}inear transformation method, \ourmethod{}.
In the \ourmethod{} CNN, partially trainable filters are configured within each Conv layer, which are referred to as \emph{seed filters} or \emph{seed Conv}.
The feature maps produced via seed filters are called \emph{seed feature maps}.
Subsequently, we split the seed feature maps into many small groups.
The remaining feature maps can be generated using diverse NLFs, each configured with different hyperparameters for each group of seed feature maps.
The hyperparameters of NLFs are fixed, ensuring transmission friendliness when cloud-assisted.
The seed feature maps and the generated feature maps are concatenated into the Conv layer's output.
The generation process of \ourmethod{} consists entirely of tensor manipulation operators and nonlinear transformation operations.
Compared with GhostNet and SineFM, there is no additional cheap Conv and Batch Normalization.
Therefore, \ourmethod{} can achieve better model training/inference acceleration effects.

Then, this paper designs the sparse \ourmethod{} Conv by reasonably setting the relationship between the seed feature map ratio and the number of Conv groups, thereby further reducing the \ourmethod{} model's parameter count and FLOPs while maintaining performance.
This paper also analyzes four types of NLFs: Sinusoidal, Monomial, Gaussian, and Laplace. It verifies across multiple datasets that the feature maps generated using the Sinusoidal function achieve the best performance.

Finally, this paper conducts extensive experiments on the CIFAR-10, CIFAR-10-C, Icons-50, ImageNet-1K, and ImageNet-C datasets, and the experimental results demonstrate that our proposed \ourmethod{} CNN models achieve strong robustness and resource-saving performance.
For example, the accuracy of \ourmethod{} ResNet-18 on Icons-50 dataset is about 2.86\% higher than that of vanilla ResNet-18. 
On the ImageNet-C dataset, the accuracy of \ourmethod{}-EfficientNet-ES is appropriately 1.1\% higher than the vanilla EfficientNet-ES.
In particular, with distributed data parallel training on ImageNet-1K, the \ourmethod{} ResNet-101 achieves a 53\% speedup over vanilla ResNet-101 while maintaining accuracy.
We also deploy \ourmethod{} models on resource-constrained devices (Raspberry Pi 4B), and on-device measurements show that the \ourmethod{} models outperform in terms of inference speed, memory usage, power, and energy consumption.

\noindent In summary, our main contributions are as follows.
\begin{itemize}

\item This paper proposes \ourmethod{}, which groups seed feature maps and uses NLFs with fixed hyperparameters to generate diverse feature maps on demand for each Conv layer. This method improves the diversity of generated feature maps and enhances the CNN model's robustness without increasing complexity.

\item This paper proposes a sparse mode of \ourmethod{}. By reasonably setting the relationship between the seed feature map ratio and the number of Conv groups, it further reduces the number of parameters and FLOPs of the \ourmethod{} CNN model while ensuring performance.

\item Extensive experiments on the CIFAR-10, CIFAR-10-C, and Icons-50, ImageNet-1K, and ImageNet-C datasets show that our proposed \ourmethod{} achieves impressive performance in terms of model robustness, training/inference speedup.

\end{itemize}

The remainder of the paper is organized as follows. Section~\ref{ref:2-related} reviews the work on cloud-assisted CNN training and deployment, and CNN. Section~\ref{ref-proposedapproach} describes the proposed \ourmethod{} method in detail. Section~\ref{ref-experiments} presents our evaluation results, and Section~\ref{ref-conclusion} concludes the paper.

\section{Related Work}\label{ref:2-related} 
This section briefly introduces cloud-device training and deployment and the Convolutional Neural Network (CNN) models that inspire our work.
\subsection{Cloud-Device Training and Deployment}
Due to the limited resources of IoT devices, cloud-device collaborative training and deployment of CNN models have received widespread attention.
The existing architectures can be classified into three categories:

\emph{Cloud-Only}. The cloud-only architecture trains and deploys the CNN models on the cloud server.
The user sends data and service requests to the cloud server, and after the cloud server processes the data, it sends the results to the users~\cite{Hazelwood@Applied, Georganas@Anatomy, Jiang@Chameleon, Liu@A}.
For example, Liu \emph{et al.}~\cite{Liu@A} proposed to deploy the CNN model on the cloud server, and the device obtains recognition services by uploading the collected data to the cloud server.
They also process data on edge servers~\cite{He@Providing, Ding@Resource, Abbas@Mobile, Wang@Agile, Ding@A, Wang@a} to reduce the amount of data uploaded to the cloud.
However, the quality of services of cloud-only architecture is fully constrained by the network conditions.

\emph{Cloud-Device Collaboration}. The second architecture splits the CNN model into two parts: partial CNN layers are deployed on the cloud server, and other parts on the IoT devices.
The cloud and IoT devices are collaborative training and inference of the split CNN model~\cite{Kang@Neurosurgeon, Li@JALAD, Stefanos@SPINN}.
For example, Kang \emph{et al.}~\cite{Kang@Neurosurgeon} proposed to divide the CNN into a head running on the device and a tail running on the cloud, and decide the split point based on the load of the device and the conditions of the cloud and model.
Stefanos \emph{et al.}~\cite{Stefanos@SPINN} proposed a cloud-device collaborative inference method for the cloud and device with compression technologies to reduce the parameter exchange.
However, cloud-device collaborative methods have to search for the optimal partitioning points labor-intensively for different CNNs.

\emph{Cloud-Assisted Training and Device Deployment}. The third architecture is that the cloud server first trains the CNN model and then sends the trained CNN model to the IoT device for deployment~\cite{Ding@Towards, Yu@Localized, xie2024multivision, Teerapittayanon@BranchyNet, Lu@TFormer, xie25nestquant}.
We refer to it as the \emph{Cloud-Assisted Architecture}.
For example, Ding \emph{et al.}~\cite{Ding@Towards} proposed to train a CNN model containing a few learnable parameters on the cloud server and then send the trained CNN model to the IoT device for deployment.
Lu \emph{et al.}~\cite{Lu@TFormer} proposed to train a lightweight vision transformer model on the cloud server, and then send the trained vision transformer model to the IoT devices for deployment.
Furthermore, they reduce the number of model parameters transmitted from the cloud to IoT devices by replacing the attention mechanism with multi-scale pooling technology.

The first and second architectures are closely related to network conditions.
When the network connection is unstable or unavailable, IoT devices will provide poor-quality services or even be unable to provide services.
The proposed \ourmethod{} is based on the cloud-assisted training and IoT device deployment architecture.

\subsection{Convolutional Neural Network Models}~\label{ref: cnn}
Because of the convolution operation's excellent feature-extraction capabilities, many popular CNN models have been proposed.
For example, Simonyan \emph{et al.}~\cite{Karen@Very} proposed the VGG model through stacked convolution operations with multiple model versions in different layers.
He \emph{et al.}~\cite{resnet} proposed the ResNet with residual branching, which solves the problem of degradation caused by deepening the convolutional layers.
Ding \emph{et al.}~\cite{Ding_2021_CVPR} introduced the re-parameterization technology to speed up the inference of the CNN model.
To deploy CNN models in resource-constrained IoT devices, many lightweight CNN models have been proposed, such as MobileNet~\cite{Howard@MobileNets,mobilenetv2, Howard@Searching}, EfficientNet~\cite{tan19efficientnet, efficientnet20lite, efficientnet19edgetpu, tan21efficientnetv2}, ShuffleNet~\cite{Zhang@ShuffleNet, Ma@ShuffleNet}, and GhostNet~\cite{han2020ghostnet}, etc.
Some methods also consider the transmission cost and robustness of CNNs, such as SineFM~\cite{Lu@Seed} and MonoCNN~\cite{Ding@Towards} for cloud-device collaborative deployment.

\begin{table}[ht]
\centering
\caption{Comparison with the Existing Convolutions with Conv($n$,$m$,$k$,$g$), height is $h$, width is $w$, and $\xi=\mathrm{Gcd}(n,\frac{n}{r})$. \label{tab:speed}}
\resizebox{.5\textwidth}{!}{%
\begin{tabular}{clcc}
\toprule
\multirow{1}*{\makecell[c]{Method}}&
\multicolumn{1}{c}{\multirow{1}*{\makecell[c]{Neural Network Modules}}}&
\multirow{1}*{\makecell[c]{\#Ops}}&
\multirow{1}*{\makecell[c]{\#FLOPs}}\\
\midrule
Vanilla & {\color{blue}Conv($n$,$m$,$k$,1)} & 1 & $whmnk^2$\\
Mono & {\color{blue}Conv($n$,$m$,$k$,1)} & 1 & $whmnk^2$ \\
Ghost & {\color{blue}Conv($n$,$\frac{m}{r}$,$k$,1)}, {\color{red}Conv($\frac{m}{r}$,$\frac{m(r\!-\!1)}{r}$,$d$,$\frac{m}{r}$)} & 2 & $wh\frac{m}{r}(nk^2\!+\!(r\!-\!1)d^2)$ \\
SineFM & {\color{blue}Conv($n$,$\frac{m}{r}$,$k$,1)}, {\color{red}Conv($\frac{tm}{r}$,$\frac{m(r\!-\!1)}{r}$,1,$\frac{m}{r}$)}, {\color{orange}BN($\frac{m}{r}$)$\times t$} & 2+$t$ &  $wh\frac{m}{r}(nk^2\!+\!t(r\!+\!1))$  \\ 
\cellcolor{gray!20}GroupNL & \cellcolor{gray!20}{\color{blue}Conv($n$,$\frac{m}{r}$,$k$,1)} & \cellcolor{gray!20}1 & \cellcolor{gray!20}$wh\frac{m}{r}nk^2$ \\ \midrule
Depthwise & {\color{blue}Conv($n$,$n$,$k$,$n$)} & 1 & $whn^2k^2\frac{1}{n}$\\
\cellcolor{gray!20}\makecell[c]{GroupNL (Sp)} & \cellcolor{gray!20}{\color{blue}Conv($n$,$\frac{n}{r}$,$k$,$\xi$)} & \cellcolor{gray!20}1 & \cellcolor{gray!20}$whn^2k^2\frac{1}{r\xi}$ \\
\bottomrule
\end{tabular}%
}
\end{table}

The most similar methods are Ghost-like Conv, i.e., GhostNet~\cite{han2020ghostnet} and SineFM~\cite{Lu@Seed}, 
which uses a \textbf{seed Conv} to get partial feature maps and leverage a transformation to generate the rest.

\begin{itemize}
    \item The GhostNet generates the feature maps with cheap Conv Transformation~\cite{han2020ghostnet}.
    The ``cheap Conv'' is a variant of depthwise Conv~\cite{Howard@MobileNets}, in which the number of output channels is a multiple of the number of input channels, and the groups of cheap Conv are equal to the common factor.
    \item The SineFM mixes nonlinear functions, Batch Normalization (BN), and cheap Conv for its feature maps generation.
    It performs better than GhostNet via nonlinearities; however, the introduced BN and cheap Conv result in slower computational speed.
\end{itemize}

These Ghost-like Conv methods cannot achieve better acceleration because they use many \textbf{Neural Network Modules, e.g., cheap Conv \& BN}.

As shown in Table~\ref{tab:speed}, for a comparison, we propose GroupNL Conv to solve this issue without using cheap Conv and BN, which only uses a \textbf{seed Conv} with some lightweight \textbf{Tensor Manipulation Operators} like $\tt{torch.cat}$, $\tt{torch.split}$, and $\tt{torch.repeat}$, achieving a better accuracy and speed.

\section{The Proposed Method: \ourmethod{}} \label{ref-proposedapproach}

We introduce \ourmethod{}, a low-resource and robust CNN design method for cloud-assisted IoT deployment, where models are trained in the cloud and deployed on resource-constrained devices.
The \ourmethod{} method addresses two key challenges: the demand for real-life resource-efficient training/inference, which prior lightweight alternatives have not fully met, and the need for robustness against real-world corruptions like adverse weather.

We describe the proposed \ourmethod{} in detail as follows: (i) preliminaries in Section~\ref{ref-preliminaries}, (ii) detailed design of \ourmethod{} in Section~\ref{ref-groupnl}, and (iii) training and inference acceleration analysis of \ourmethod{} in Section~\ref{ref-resource-analysis}.

\subsection{Preliminaries} \label{ref-preliminaries}

We illustrate our motivation by analyzing the most relevant works, the nonlinear transformation method MonoCNN~\cite{Ding@Towards}, and Ghost-like Conv, including GhostNet~\cite{han2020ghostnet} and SineFM~\cite{Lu@Seed}, respectively.

Given the input feature maps ${\bm{x}} \!\in\! \mathbb{R}^{h\times w\times c_\mathrm{in}}$ and output feature maps ${\bm{y}} \!\in\! \mathbb{R}^{h^{\!'\!}\times w^{\!'\!}\times c_\mathrm{out}}$ of this Conv layer, where the $h$ and $w$ are the spatial dimensions of input feature maps, and $h^{'}$ and $w^{'}$ for output feature maps.
$c_\mathrm{in}$, $c_\mathrm{out}$ are the number of input and output channels, respectively.

For a detailed description, we introduce two concepts:
(i) {\emph{Seed filters}} refer to a small number of Conv filters, or referred to as \emph{seed Conv};
(ii) {\emph{Seed feature maps}} refer to feature maps generated by seed filters via seed Conv operation.

We denote seed filters as ${\bm{W}}_\mathrm{seed} \!\in\! \mathbb{R}^{k\times k \times c_\mathrm{in}\times c_\mathrm{seed}}$, where $k$ is the kernel size.
The number of seed filters is $c_\mathrm{seed}$, and the number of generated filters is $c_\mathrm{gen}$, where $c_\mathrm{seed}\!+\!c_\mathrm{gen}\!=\!c_\mathrm{out}$.
We denote an integer reduction ratio $r$ representing the relationship between $c_\mathrm{seed}$ and $c_\mathrm{out}$, i.e., $r={c_\mathrm{out}}/{c_\mathrm{seed}}$.

We also set the trainable $\tt{torch.nn.Module}$ operations in bold in equation, e.g., $\mathrm{\bf{Conv}}$ and $\mathrm{\bf{BN}}$.

{\emph{MonoCNN}}~\cite{Ding@Towards}. For a Conv layer, some filters are designated as seed filters, and the remaining filters are generated based on seed filters and one data-agnostic nonlinear transformation function (NLF). 
Formally, given seed filters and a monomial NLF $\Psi_\mathrm{mono} (\cdot)$, the MonoConv obtains the remaining filters ${\bm{W}}_\mathrm{gen} \!\in\! \mathbb{R}^{k\times k \times c_\mathrm{in}\times c_\mathrm{gen}}$ as follows:
\begin{equation}
\begin{array}{l}
{\bm{W}}_\mathrm{gen} = \Psi_\mathrm{mono} ({\bm{W}}_\mathrm{seed})
\end{array},
\label{eq:mono_nonlinear}
\end{equation}
Then, concatenating the seed filter ${\bm{W}}_\mathrm{seed}$ and the generated filters ${\bm{W}}_\mathrm{gen}$ to obtain the filters  ${\bm{W}}\!\in\!\mathbb{R}^{k\times k \times c_\mathrm{in}\times c_\mathrm{out}}$:
\begin{equation}
\begin{array}{l}
{\bm{W}} = \mathrm{Concat}[{\bm{W}}_\mathrm{seed}; {\bm{W}}_\mathrm{gen}]
\end{array},
\label{eq:mono_weight}
\end{equation}
Finally, perform a vanilla Conv operation on input feature maps to obtain the output ${\bm{y}}$ of MonoConv:
\begin{equation}
\begin{aligned}
{\bm{y}} 
= {\bm{W}}* {\bm{x}}
= \mathrm{Concat}[{\bm{W}}_\mathrm{seed}; \Psi_\mathrm{mono} ({\bm{W}}_\mathrm{seed})] * {\bm{x}} 
\ ,
\end{aligned}
\label{eq:convop}
\end{equation}
where $*$ is the Conv operation.
The MonoCNN improves the robustness of the CNN model via data-agnostic NLFs for filters, but still with a complete Conv operation.

{\emph{GhostNet}}~\cite{han2020ghostnet}. The Ghost Conv introduces $k \times k$ seed Conv to generate seed feature maps:
\begin{equation}
\begin{array}{l}
{\bm{y}}_\mathrm{seed} = {\bm{W}}_\mathrm{seed}* {\bm{x}}
=\mathop{\mathrm{\bf{Conv}}}\limits_{\ k\times k}({\bm{x}})
\end{array},
\label{eq:seed_weight}
\end{equation}
where ${\bm{y}}_\mathrm{seed} \!\in\! \mathbb{R}^{h^{\!'\!}\times {w^{\!'\!}}\times c_\mathrm{seed}}$.
For the remaining feature maps, GhostNet introduces additional cheap linear transformation, i.e., the $d\!\times\! d$ cheap Conv to process the ${\bm{y}}_\mathrm{seed}$ for generating features:
\begin{equation}
\begin{aligned}
{\bm{y}}= \mathrm{Concat}[{\bm{y}}_\mathrm{seed}; {\bm{y}}_\mathrm{gen}]
= \mathrm{Concat}[{\bm{y}}_\mathrm{seed}; \mathop{\mathrm{\bf{Conv}}}\limits_{d\times d}({\bm{y}}_\mathrm{seed})] \ .
\end{aligned}
\label{eq:ghostnet}
\end{equation}

However, the cheap Conv introduced by GhostNet slows down training and inference. 
Also, the parameters of cheap Conv are trainable, so GhostNet's robustness is completely restricted to the training data.

{\emph{SineFM}}~\cite{Lu@Seed}. For a SineFM Conv, similar to the Ghost Conv, given seed filters ${\bm{W}}_\mathrm{seed}$ and input feature maps ${\bm{x}}$, we can obtain seed feature maps ${\bm{y}}_\mathrm{seed}$ by the Eq.~\ref{eq:seed_weight}.
Then, $t$-NLFs $\bm{{\Psi}}_\mathrm{sinefm} \!=\! {\{\Psi_{i} \| i\!=\!1,\!\dots\!,t\}}$ process the seed feature maps to obtain expanded feature maps ${\bm{y}}_{\mathrm{exp}}\!\in\!\mathbb{R}^{h^{\!'\!}\!\times\! w^{\!'\!}\!\times\! t\cdot c_\mathrm{seed}}$ with more features:
\begin{equation}
\begin{array}{l}
{\bm{y}}_\mathrm{exp} = \mathrm{Concat}[{\bm{y}}_{\mathrm{exp}_1}; \cdots; {\bm{y}}_{\mathrm{exp}_t}]= \mathop{\mathrm{Concat}}\limits_{i= 1,\cdots,t}[{\bm{y}}_{\mathrm{exp}_i}]\ ,\\
{\bm{y}}_{\mathrm{exp}_i} = \mathrm{\bf{BN}}_i(\Psi_i({\bm{y}}_\mathrm{seed})) \ ,
\end{array}
\label{eq:expand_sinefm}
\end{equation}
where ${\bm{y}}_{\mathrm{exp}_i}\!\in\!\mathbb{R}^{h^{\!'\!}\!\times\!w^{\!'\!}\!\times\!c_\mathrm{seed}}$, and $i\!=\!1,\!\cdots\!,\!t$.
The $t$ is the time of expanding seed feature maps.
The expanded feature maps will exceed the number of feature maps required for this layer.
Each expanded seed feature maps ${\bm{y}}_{\mathrm{exp}_i}$ also process a specific $\mathrm{\bf{BN}}_i$.
Then, expanded feature maps are processed by $1\!\times\! 1$ cheap Conv to align the $c_\mathrm{gen}$ dimensions:
\begin{equation}
\begin{array}{l}
{\bm{y}}_\mathrm{gen} = \mathop{\mathrm{\bf{Conv}}}\limits_{1\times 1}({\bm{y}}_{\mathrm{exp}}) \ .
\end{array}
\label{eq:expand_gen}
\end{equation}

Finally, the generated feature maps are concatenated with the seed feature maps to form the output of the SineFM layer as follows:
\begin{equation}
\begin{aligned}
{\bm{y}} &= \mathrm{Concat}[{\bm{y}}_\mathrm{seed}; {\bm{y}}_\mathrm{gen}] \\
&= \mathrm{Concat}[{\bm{y}}_{\mathrm{seed}}; \mathop{\mathrm{\bf{Conv}}}\limits_{1\times 1}(\mathop{\mathrm{Concat}}\limits_{i= 1,\cdots,t}[\mathrm{\bf{BN}}_i(\Psi_i({\bm{y}}_\mathrm{seed}))])] \ .
\end{aligned}
\label{eq:sinefm}
\end{equation}

\begin{figure}
\centering
\includegraphics[width=1.\linewidth]{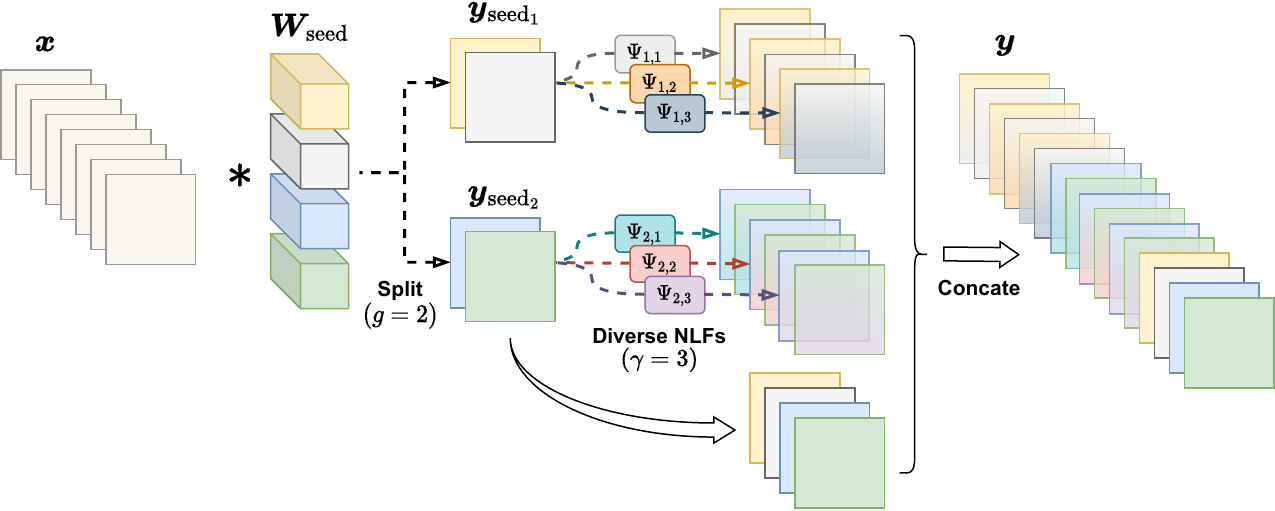}
\caption {
The example of GroupNL Conv with $c_\mathrm{in} \!=\! 8, c_\mathrm{out}\!=\! 16$, and the number of seed features is $c_\mathrm{seed} \!=\! 4$, the groups for splitting is $g\!=\!2$. 
In this case, the GroupNL can introduce $\gamma \cdot g \!=\! 6$ different NLFs for the feature maps generation.
Compared with the vanilla Conv, which needs 16 filters to compute all features, the GroupNL Conv only requires 4 filters to compute the seed features and generates the remaining features by nonlinearities and tensor manipulation operators. 
}
\label{fig:GroupNL}
\end{figure}

\subsection{Design of GroupNL}\label{ref-groupnl}

\subsubsection{Grouped Nonlinear Transformation} \label{ref-groupnl-algo}
%
The aforementioned GhostNet and SineFM methods introduce too much computation in feature maps generation by cheap Conv and BN operations, which result in slow training/inference speed.
At the same time, NLFs in SineFM are only $t$-different types.

Therefore, resource efficiency and robustness need to be further improved.
The key challenge is how to increase the \emph{diversity} of data-agnostic NLFs to improve the robustness of the model without increasing computations.

To this end, we propose the {\textbf{Group}}ed {\textbf{N}}on{\textbf{L}}inear transformation (GroupNL) method.
Formally, given the seed filters ${\bm{W}}_\mathrm{seed}$ and input feature maps ${\bm{x}}$, we can obtain some seed feature maps ${\bm{y}}_\mathrm{seed}$ with the Eq.~\ref{eq:seed_weight}.

\emph{Splitting and Grouped Algorithm}. We split seed feature maps into $g$ groups in $c_\mathrm{seed}$ dimension instead of expanding:
\begin{equation}
\begin{array}{l}
{\bm{y}}_\mathrm{seed} = \mathrm{Concat}[{\bm{y}}_{\mathrm{seed}_1}; \cdots; {\bm{y}}_{\mathrm{seed}_g}] = \mathop{\mathrm{Concat}}\limits_{i= 1,\cdots,g}[{\bm{y}}_{\mathrm{seed}_i}] 
\end{array},
\label{eq:groupnl_split}
\end{equation}
where $g$ is a positive integer, where its modulo result is $\mathrm{Mod}(c_\mathrm{seed}, g)\!=\!0$, and ${\bm{y}}_{\mathrm{seed}_i}\!\in\!\mathbb{R}^{h^{\!'\!}\times {w^{\!'\!}}\times \frac{c_\mathrm{seed}}{g}}, i=1,\!\cdots\!,\!g$.
For the remaining dimension $c_\mathrm{gen} \!=\! c_\mathrm{out} \!-\! c_\mathrm{seed}$, we want to represent it by $c_\mathrm{seed}$ without expanding the dimension.

We denote a multiplier $\gamma \!=\! 
\frac{c_\mathrm{gen}}{c_\mathrm{seed}} \!=\! r \cdot\frac{c_\mathrm{out}}{c_\mathrm{in}} - 1$, so each group of seed feature maps should be copied $\gamma$ times for filling the remaining dimension.
Thus, the total number of copied groups of feature maps is $\gamma \cdot g$. 

\emph{Grouped Diverse Nonlinear Transformation}. 
By splitting and copying, we have $\gamma \cdot g$ groups of identical seed feature maps, and no additional dimensions are introduced.
We can introduce $\gamma \cdot g$ data-agnostic different nonlinear transformations, i.e., diverse NLFs $\bm{{\Psi}}_\mathrm{gnl} \!=\! {\{\Psi_{i,j} \| i\!=\!1,\dots,g; j\!=\!1,\dots,\gamma\}}$ to these identical groups.
Subsequently, for each copied group of seed feature maps, NLFs with different hyperparameter settings are used to generate corresponding feature maps:
\begin{equation}
\begin{array}{l}
{\bm{y}}_\mathrm{gen} = \mathrm{Concat}[{\bm{y}}_{\mathrm{gen}_1}; \cdots; {\bm{y}}_{\mathrm{gen}_g}] = \mathop{\mathrm{Concat}}\limits_{i= 1,\cdots,g}[{\bm{y}}_{\mathrm{gen}_i}] \ , \\
{\bm{y}}_{\mathrm{gen}_i} = \mathop{\mathrm{Concat}}\limits_{j= 1,\cdots,\gamma}[\Psi_{i,j}({\bm{y}}_{\mathrm{seed}_i})] \ ,
\end{array}
\label{eq:groupnl_nonlinear}
\end{equation}
where $i\!=\!1,\!\cdots\!,\!g$, $j\!=\!1,\!\cdots\!,\!\gamma$, ${\bm{y}}_{\mathrm{gen}_i}\!\in\!\mathbb{R}^{h^{\!'\!}\times {w^{\!'\!}}\times \gamma\cdot\frac{c_\mathrm{seed}}{g}}$, and ${\bm{y}}_{\mathrm{gen}} \!\in\! \mathbb{R}^{h^{\!'\!}\times {w^{\!'\!}}\times \gamma\cdot c_\mathrm{seed}}, \gamma\cdot c_\mathrm{seed} = c_\mathrm{gen}$.

As shown in Fig.~\ref{fig:GroupNL}, we can describe the operations of grouped diverse nonlinear transformation in detail with an example of $c_\mathrm{in} \!=\! 8, c_\mathrm{out} \!=\! 16, c_\mathrm{seed} \!=\! 4, g\!=\!2$.
In this case, the number of generated features $c_\mathrm{gen} \!=\! 16\!-\! 4 \!=\! 12$, the multiplier $\gamma \!=\! c_\mathrm{gen} / c_\mathrm{seed} \!=\! 12 / 4 \!=\! 3$, and introducing $|\bm{{\Psi}}_\mathrm{gnl}|\!=\!\gamma \cdot g \!=\! 6$ different NLFs for feature maps generation.

\emph{GroupNL Conv}. Finally, generated feature maps are concatenated with the seed feature maps to produce the output of the standard GroupNL Conv layer, as follows:
\begin{equation}
\begin{aligned}
{\bm{y}} &= \mathrm{Concat}[{\bm{y}}_{\mathrm{seed}}; {\bm{y}}_{\mathrm{gen}}]\\
&=\mathrm{Concat}[{\bm{y}}_{\mathrm{seed}}; \mathop{\mathrm{Concat}}\limits_{i= 1,\cdots,g}[\mathop{\mathrm{Concat}}\limits_{j= 1,\cdots,\gamma}[\Psi_{i,j}({\bm{y}}_{\mathrm{seed}_i})]]] \ .
\end{aligned}
\label{eq:groupnl}
\end{equation}

\emph{Sparse GroupNL Conv}. In addition to the standard GroupNL Conv, we introduce a more lightweight module, termed \emph{Sparse GroupNL Conv}.
In this design, the seed filters ${\bm{W}}_\mathrm{seed}$ are grouped for thinning, where the number of groups $\xi$ in seed Conv is set to the Greatest Common Divisor (GCD) between $c_\mathrm{in}$ and $c_\mathrm{seed}$, i.e., $\xi\!=\!\mathrm{Gcd}(c_\mathrm{in}, c_\mathrm{seed})$.
The number of parameters, FLOPs, and energy consumption in Sparse GroupNL Conv are close to or slightly lower than depthwise Conv, while offering higher computational efficiency, as profiling in Section~\ref{sec:module_profiling}.

Therefore, Sparse GroupNL Conv can be used to replace depthwise Conv in lightweight models, e.g., MobileNet, EfficientNet, to improve the robustness.

Meanwhile, we observe that Sparse GroupNL Conv preserves performance in large-scale CNNs with bottleneck structures (e.g., ResNet-101), whereas the standard GroupNL Conv is still required for compact models based on basic blocks (e.g., ResNet-18/34).

The PyTorch-like pseudocode of the GroupNL Conv layer is demonstrated in Algorithm~\ref{alg:code}.
As shown, its generating procedures are all \emph{tensor manipulation operators}, e.g., splitting $\tt{torch.split}$, concatenation $\tt{torch.cat}$, copying $\tt{torch.repeat}$ and $\tt{torch.repeat\_interleave}$, and data-agnostic and random \emph{nonlinear transformation} operations, without any extra $\tt{torch.nn.Module}$ (Conv and BN) compared with GhostNet in Eq.~\ref{eq:ghostnet} and SineFM in Eq.~\ref{eq:sinefm}.

\begin{algorithm}[t]
\caption{GroupNL Conv: PyTorch Pseudocode\label{alg:code}}
\definecolor{codeblue}{rgb}{0.25,0.5,0.5}
\definecolor{codekw}{rgb}{0.85, 0.18, 0.50}
\begin{lstlisting}[language=python, mathescape]
# func: nonlinear transformation function
# sparse: is seed filter sparse
# g: No. of nonlinear transformation groups 
# C_i, C_s, C_o: No. of input/seed/output features
class GroupNLConv2d(nn.Conv2d):
    def __init__(self, C_in, C_out, C_s, g, func, sparse, **kwargs):
        super(GroupNLConv2d, self).__init__(
            C_in, C_out, **kwargs)
        self.weight = None  # ensure non-learnable
        self.sparse = sparse # sparsity flag
        # No. of generated features
        C_g = C_s * (ceil(C_o/C_s) - 1)
        # seed filters for generating features
        self.conv = nn.Conv2d(C_i, C_s, \
            groups=math.gcd(C_i, C_s) \
            if self.sparse else 1, **kwargs)
        # No. of grouped seed features lists
        groups_s = [C_s//g for _ in range(g)]
        gamma = C_g/C_s # multiplier gamma
        # diverse random hyperparameters of func
        self.hy=nn.Parameter(torch.rand(gamma*g))
        # ensure non-learnable hyperparameters
        self.hy.requires_grad = False
        
    def forward(self, x):
        # feature maps shape [B, C, H, W]
        y_s = self.conv(x) # seed features
        # split seed features into groups
        y_sp = torch.split(y_s, groups_s)
        # copy gamma times feature maps
        y_cp = torch.cat([y_sp[j].repeat(
            [1, gamma, 1, 1]) for j in range(g)]) 
        # copy hyperparameters of func
        hy = self.hy.repeat_interleave(C_s/g).reshape([1, C_g, 1, 1]) 
        y_g = func(y_cp, hy) # generate C_g features
        return torch.cat([y_s, y_g])
\end{lstlisting}
\end{algorithm}

%
\subsubsection{Diversities of Nonlinearities and Feature Maps}
The number of NLFs in GroupNL is $|\Psi_\mathrm{gnl}| \!=\! \gamma \cdot g$, which is usually larger than the fixed number of $t$ NLFs in SineFM $|\Psi_\mathrm{sinefm}| \!=\! t$, which act on all seed feature maps to expand or limit feature diversity.

\emph{Diversities of Data-agnostic NLFs}. Assuming there are $64$ seed feature maps, and the number of output channels is $c_\mathrm{out}\!=\!256$, we can use different NLFs to generate the remaining $192$ feature maps.

For SineFM, assuming $t\!=\!5$, it needs to generate $t\cdot c_\mathrm{seed} \!=\! 5\cdot 64\!=\! 320$ expanding features and utilizes a cheap Conv to align them into $192$ features.
So, the number of NLFs is $5$.

For \ourmethod{}, the $\gamma \!=\!  c_\mathrm{gen}/c_\mathrm{seed}\!=\!192/64\!=\!3$.
For $g=2$, the number of different NLFs is $|\Psi_\mathrm{gnl}| \!=\! \gamma \cdot g \!=\! 3 \cdot 2 \!=\! 6$; for $g\!=\!4$ is $|\Psi_\mathrm{gnl}| \!=\! \gamma \cdot g \!=\! 3 \cdot 4 \!=\!12$; for $g\!=\!8$ is $|\Psi_\mathrm{gnl}| \!=\! \gamma \cdot g \!=\! 3 \cdot 8 \!=\!24$; and for the fully grouped $g\!=\!64$ is $|\Psi_\mathrm{gnl}| \!=\! \gamma \cdot g \!=\! 3 \cdot 64 \!=\! 192 \!\gg\! |\Psi_\mathrm{sinefm}| \!=\! 5$.

Compared with the SineFM, the diversity of NLFs is significantly improved without increasing the number of feature maps in \ourmethod{}.

\begin{figure}
\centering
\includegraphics[width=.9\linewidth]{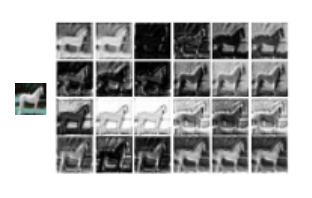}
\caption {The visualization of feature maps generated based on the diverse NLFs with 24 different hyperparameter settings. 
The diversity of the generated feature maps is greatly improved while maintaining the similarity.
}
\label{fig:horse_fm}
\end{figure}

\emph{Diversities of Generated Feature Maps}. We can visualize a given input image processing with 24-different NLFs as an example, as shown in Fig.~\ref{fig:horse_fm}. 
The richness of the generated feature maps is greatly improved while maintaining the similarity between different features.

\subsection{Training \& Inference Acceleration Analysis}\label{ref-resource-analysis} 
\emph{Cloud Training Acceleration}.
For cloud-assisted CNN models, the multi-GPU training on cloud servers is mainly based on data parallelism.
In practical multi-GPU training, such as Data Parallelism (DP)~\cite{Paszke19pytorch} and Distributed Data Parallelism (DDP)~\cite{li2020pytorchddp}, the bandwidth of inter-GPU memory data transfers is an important constraint on the training speed compared to the powerful and fast computation kernels.

So, the number of gradients can be further optimized by reducing inter-GPU scatter and by optimizing different data parallelism levels.
Assuming each trainable parameter in a CNN model maintains a gradient, the number of gradients $\mathcal{G}$ corresponds to the number of trainable parameters $\mathcal{P}$.
For $N$-GPUs DP, one iteration data communication in the main GPU is $(N\!-\!1)\mathcal{G}$; for the rest $N-1$ GPUs are $(N\!-\!1)\mathcal{G}$, where each GPU sends $\mathcal{G}$.
For $N$-GPUs DDP, one iteration data communication in a single GPU is $\frac{2(N-1)}{N}\mathcal{G}$.
Similar to lightweight models, GroupNL essentially achieves training acceleration by reducing the $\mathcal{P}$ (and thus the $\mathcal{G}$).

Based on this premise, we analyze the theoretical gradient exchange in multi-GPUs memory of vanilla CNN, GhostNet, SineFM, and GroupNL in training.

Ignoring the bias, for a vanilla Conv layer, the number of parameters and gradients can be described as follows:
\begin{equation}
\begin{array}{l}
{\mathcal{P}_\mathrm{conv} = \mathcal{G}}_\mathrm{conv} = c_\mathrm{in}c_\mathrm{out}k^2 \ .
\end{array}
\label{eq:cnn_grad}
\end{equation}

For a Ghost Conv layer, the number of parameters and gradients from the seed Conv and $d\!\times\!d$ cheap Conv are:
\begin{equation}
\begin{aligned}
{\mathcal{P}_\mathrm{ghost} = \mathcal{G}}_\mathrm{ghost} &= c_\mathrm{in}c_\mathrm{seed}k^2 + c_\mathrm{seed}c_\mathrm{gen}d^2\frac{1}{c_\mathrm{seed}}\\
&=c_\mathrm{in}c_\mathrm{out}k^2\frac{1}{r} + c_\mathrm{out}\frac{r\!-\!1}{r}d^2 \ ,
\end{aligned}
\label{eq:ghostnet_grad}
\end{equation}
\noindent and the number of parameters and gradients for the SineFM from the seed Conv, $1\!\times\!1$ cheap Conv, and BN, are:
\begin{equation}
\begin{aligned}
{\mathcal{P}_\mathrm{sinefm} =\mathcal{G}}_\mathrm{sinefm} &= c_\mathrm{in}c_\mathrm{seed}k^2 + tc_\mathrm{seed}c_\mathrm{gen}\frac{1}{c_\mathrm{seed}} + 2tc_\mathrm{seed}\\
&= c_\mathrm{in}c_\mathrm{out}k^2\frac{1}{r} + c_\mathrm{out}\frac{t(r\!+\!1)}{r} \ .
\end{aligned}
\label{eq:ghostnet_grad}
\end{equation}
The proposed GroupNL Conv layer only contains the learnable standard seed filters and learnable sparse seed filters, assuming $\xi=\mathrm{Gcd}(c_\mathrm{in}, c_\mathrm{seed})$, and the number of parameters and gradients are shown in Eq.~\ref{eq:groupnl_grad}:

\begin{equation}
\begin{aligned}
{\mathcal{P}_\mathrm{gnl} = \mathcal{G}}_\mathrm{gnl} &= {c_\mathrm{in}c_\mathrm{seed}k^2} =  {c_\mathrm{in}c_\mathrm{out}k^2\frac{1}{r}}\ ,\\
\mathcal{P}_\mathrm{gnl}^\mathrm{sp} =\mathcal{G}_\mathrm{gnl}^\mathrm{sp} 
&= {c_\mathrm{in}c_\mathrm{seed}k^2}\frac{1}{\xi} =  c_\mathrm{in}c_\mathrm{out}k^2\frac{1}{ r\xi}\ .\\
\end{aligned}
\label{eq:groupnl_grad}
\end{equation}

Noticeably, the relation is $\mathcal{G}_\mathrm{gnl}^\mathrm{sp} \!\leq\! \mathcal{G}_\mathrm{gnl} \!=\!c_\mathrm{in}c_\mathrm{out}k^2\frac{1}{r} \!<\! \mathrm{min}\{{\mathcal{G}}_\mathrm{conv}, {\mathcal{G}}_\mathrm{ghost}, {\mathcal{G}}_\mathrm{sinefm}\}$, i.e., the number of gradients for optimization in GroupNL is minimal compared to the rest CNN design methods of ${\mathcal{G}}_\mathrm{conv}$, ${\mathcal{G}}_\mathrm{ghost}$, and  ${\mathcal{G}}_\mathrm{sinefm}$.
The sparse GroupNL also significantly reduces the gradient exchange by a factor of ${ (r\xi)}^{-1}$ compared to the ${\mathcal{G}}_\mathrm{conv}$. 

Notably, it seems that Ghost Conv and SineFM also have a small number of gradients with ${{\mathcal{G}}_\mathrm{ghost} \!<\! \mathcal{G}}_\mathrm{conv}$ and ${\mathcal{G}}_\mathrm{sinefm} \!<\! {\mathcal{G}}_\mathrm{conv}$, can also accelerate training in theory.
However, the evaluation in Table~\ref{Table:trainingAccelerate_all_cnn} shows that the GhostNet and SineFM are even slower than the vanilla CNN.
For this phenomenon, we consider this to be due to the introduction of more computational nodes involved in the backward updating, such as cheap Conv and BN, which leads to slower training instead.
The GroupNL, on the other hand, does not introduce new computational nodes through tensor manipulation operations such as $\tt{torch.cat}$, $\tt{torch.split}$, $\tt{torch.repeat}$ and $\tt{torch.repeat\_interleave}$, thus also ensures acceleration of training.

\emph{On-Device Inference Efficiency}.
For a fixed feature map size $wh$, the computational cost in FLOPs is proportional to the number of parameters, resulting in $\mathcal{F}_\mathrm{gnl}^\mathrm{sp} \!\leq\! \mathcal{F}_\mathrm{gnl} \!=\!c_\mathrm{in}c_\mathrm{out}k^2\frac{1}{r} \!<\! \mathrm{min}\{{\mathcal{F}}_\mathrm{conv}, {\mathcal{F}}_\mathrm{ghost}, {\mathcal{F}}_\mathrm{sinefm}\}$, confirming that GroupNL Conv requires significantly fewer computations than other Convs.
Furthermore, GroupNL employs the fewest $\tt{torch.nn.Module}$, contributing to its theoretically superior inference efficiency on devices.
The module-level profiling in Section~\ref{sec:module_profiling} also demonstrates the practical inference efficiency.

\subsection{Discussion} 
As the aforementioned analysis, compared with the GhostNet~\cite{han2020ghostnet} and SineFM~\cite{Lu@Seed}, the proposed \ourmethod{}-based CNN design method has two highlights:

(i) More diverse feature maps.
Given the seed filters and the number of output channels of the layer, by grouping the seed filters, more different hyperparameter settings of data-agnostic NLFs can be used to generate diverse features.

(ii) More efficient training. \ourmethod{} generates the number of feature maps without extra computational nodes like cheap Conv and BN, effectively reducing the number of gradients exchanging in the multi-GPUs training.

\section{Experiments} \label{ref-experiments}
This section introduces our experimental setup, including the datasets, baselines, and implementation details.
Then, we empirically compare accuracy, network traffic, FLOPs, and training speed on multiple benchmarks.
 
\subsection{Experimental Setup}

\subsubsection{Datasets} 
We evaluate the effectiveness of the proposed method using five widely used datasets.
CIFAR-10 and ImageNet-1K are standard datasets.
CIFAR-10-C, Icons-50, and ImageNet-C are used as corrupted datasets.

\emph{CIFAR-10}~\cite{Alex@Learning} is a multi-class natural object dataset for image classification. 
It consists of 50,000 training images and 10,000 test images in 10 categories, each with a resolution of $32\!\times\!32$ pixels.

\emph{CIFAR-10-C}~\cite{Hendrycks@Benchmarking} is a test dataset after using synthetic common perturbations and noise corruptions on the CIFAR-10 test set. It consists of 10,000 test images of 19 types of damage in 4 categories. The resolution of each image is $32\!\times\!32$ pixels.

\emph{Icons-50}~\cite{Hendrycks@Benchmarking} consists of 10,000 images in 50 categories collected from different companies. 
We set the resolution of all images to $32\!\times\!32$ pixels. 
For training, the data of one company is retained as test data, and the data of the rest of the companies is used as training data.

Please refer to~\cite{Ding@Towards} for more explanation of the CIFAR-10-C and Icons-50 datasets.

\emph{ImageNet-1K}~\cite{deng2009imagenet} is a large-scale classification dataset to verify computer vision performance. 
ImageNet-1K is a subset of the ILSVRC ImageNet, which contains 1.28 billion training images and 50K validation images from 1K different classes.
The resolution of each image is $224\!\times\!224$ pixels.

\begin{figure}
\centering
\includegraphics[width=1.\linewidth]{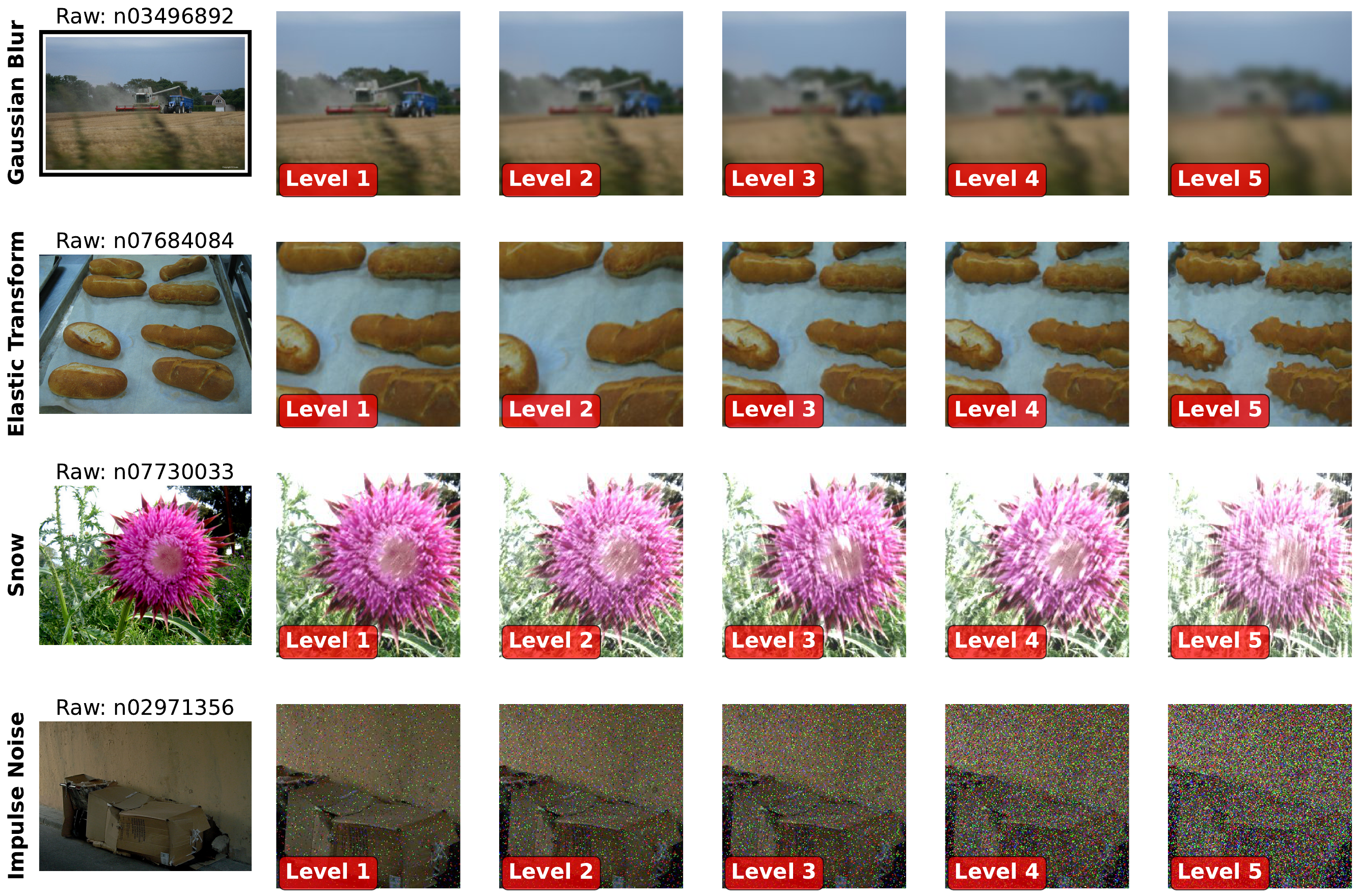}
\caption {
Visualization examples of different severity levels of corruptions on ImageNet-C.
}
\label{fig:imagenetc_example}
\end{figure}

\emph{ImageNet-C}~\cite{Hendrycks@Benchmarking} is a large-scale dataset designed to evaluate the robustness of CNNs against common corruptions. 
The ImageNet-C includes 19 types of corruptions at five severity levels (as shown in Figure~\ref{fig:imagenetc_example}) in 5 categories: \emph{Noise}, \emph{Blur}, \emph{Weather}, \emph{Digital}, and \emph{Extra}, simulating real-world challenges, e.g., environmental conditions, based on ImageNet-1K validation data.
The 19 types include: \emph{gaussian noise}, \emph{shot noise}, \emph{impulse noise}, \emph{defocus blur}, \emph{glass blur}, \emph{motion blur}, \emph{zoom blur}, \emph{snow}, \emph{frost}, \emph{fog}, \emph{brightness}, \emph{contrast}, \emph{elastic transformation}, \emph{pixelate}, \emph{JPEG compression}, \emph{speckle noise}, \emph{gaussian blur}, \emph{spatter}, and \emph{saturate}.
The resolution of the corrupted image is $224\!\times\!224$.

\subsubsection{Baseline CNNs}  
We compare the following robust and feature maps generation CNNs discussed in Section~\ref{ref-preliminaries}:
\emph{MonoCNN}~\cite{Ding@Towards}, \emph{GhostNet}~\cite{han2020ghostnet}, and \emph{SineFM}~\cite{Lu@Seed}, and the above baseline methods are based on the standard CNN models: \emph{ResNet-18/-34/-101}~\cite{resnet} and \emph{VGG11}~\cite{Karen@Very}.
%

For evaluations of large-scale datasets, i.e., ImageNet-1K and ImageNet-C, we choose 
representative lightweight CNNs widely used in mobile edge AI, including \emph{MobileNet-V2}~\cite{mobilenetv2}, \emph{MobileNet-V3}~\cite{Howard@Searching}, \emph{EfficientNet-Lite0}~\cite{efficientnet20lite}, 
\emph{EfficientNet-EdgeTPU-Small}~\cite{efficientnet19edgetpu} (\emph{EfficientNet-ES}), \emph{EfficientNet-V2-Small}~\cite{tan21efficientnetv2}.

To make the effect of training acceleration more clear, we evaluate the training speed of a larger model, i.e., ResNet-101.
For on-device resource measurements, we deploy vanilla and GroupNL-based MobileNetV3, EfficientNet-ES, ResNet-50, and ResNet-101.

\subsubsection{Implementation Details}  
The implementation details include training platforms \& libraries and experimental settings.

\textit{Training Platforms and Libraries:}
The experiments are performed on the cloud servers with NVIDIA RTX 2080Ti 11GB and RTX 4090 GPUs 24GB. 
We implement the models in PyTorch 1.12 for the NVIDIA platforms with CUDA 11.1.

\textit{Hyperparameters Settings of Baseline CNNs:}
For the hyperparameters setting, we follow the suggestion from the original papers~\cite{Ding@Towards, han2020ghostnet, Lu@Seed}.
The specific description of model hyperparameters is shown in Table~\ref{ref:hypersettings}.
\begin{itemize}
    \item {MonoCNN}~\cite{Ding@Towards} uses the monomial function as a filter generating function and samples a random and continuous exponent $\eta$ between $[1,7]$, and the exp\_factor $\bm{e}$ is $\{2,4,8,16\}$;
    \item {GhostNet}~\cite{han2020ghostnet} uses a $d\!=\!3$ grouped convolution as the cheap operation for generating feature maps.
    \item {SineFM}~\cite{Lu@Seed} uses a sinusoidal function, the ranges of period $\omega$ and shift $\phi$ are between $[1,2]$ and $[1,5]$, respectively. 
    The number of NLFs in SineFM is $t\!=\!5$.
\end{itemize}

We also set up the same scale of seed feature map numbers in GhostNet, SineFM, and \ourmethod{} with $r\!=\!2$.

\begin{figure}
\centering
\includegraphics[width=.55\linewidth]{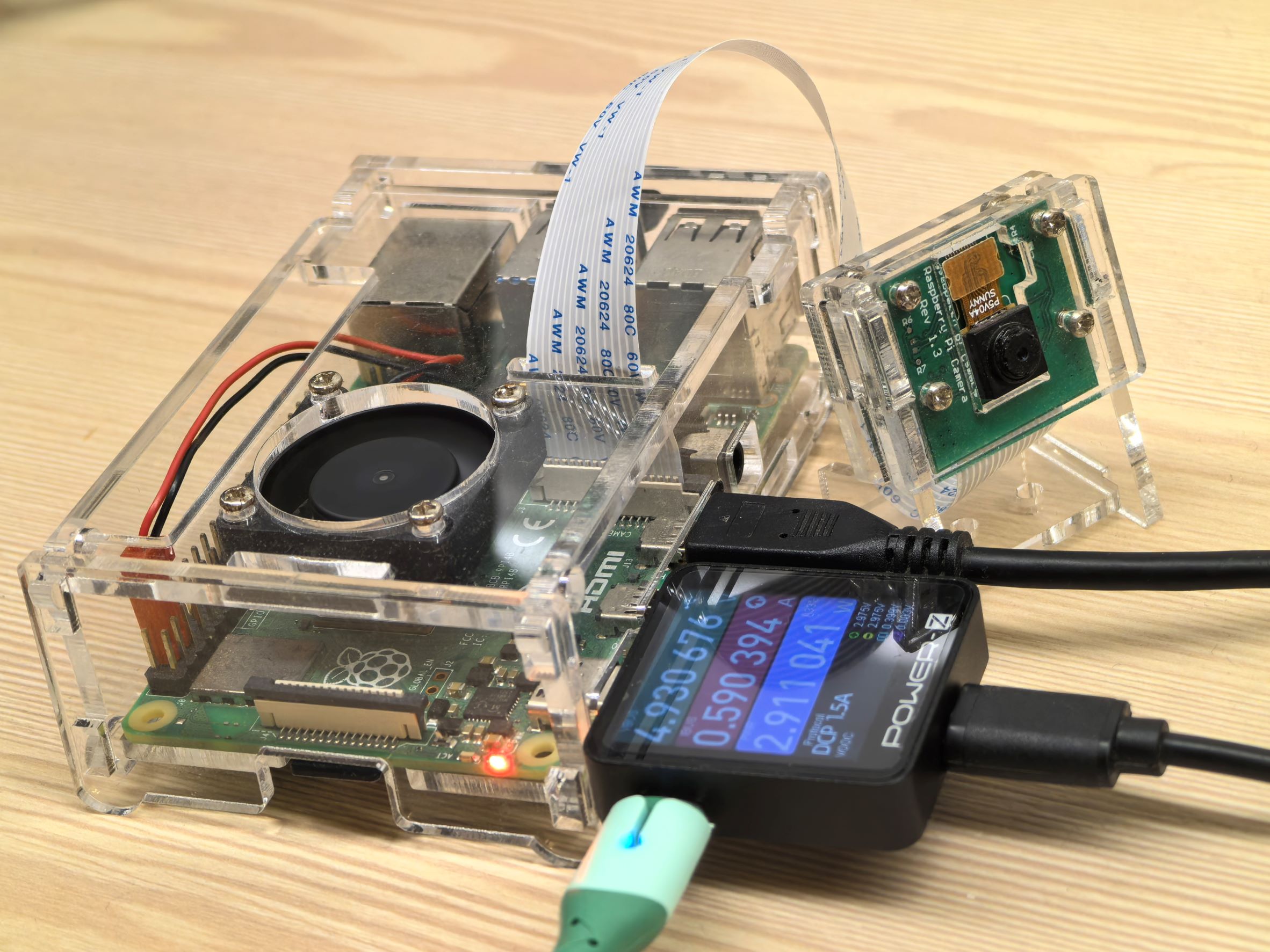}
\caption {Hardware: a Resource-constrained Raspberry Pi 4B for Deployment and a Power Meter for Energy Evaluation. 
}
\label{fig:inference_system}
\end{figure}

\textit{Experimental Settings:}
For the settings of training in CIFAR-10/Icons-50, the learning rate is 0.1, and the batch size is 128, with the SGD optimizer and cosine annealing scheduler.
For training ResNet architectures in ImageNet-1K, the epochs is 300, the learning rate is 0.001, and the batch size is 1024, with the Adam optimizer and cosine annealing scheduler.
For training lightweight CNNs in ImageNet-1K, the learning rate is 0.064, and the batch size is 1024, with the RMSprop optimizer and step scheduler. For MobileNet-V2, EfficientNet-Lite0, EfficientNet-ES, and EfficientNet-V2-S, the epoch is 450; for MobileNet-V3, the epoch is 600. 
In addition, the training evaluation is compared for the larger model, i.e., ResNet-101, in different parallel technologies in PyTorch~\cite{Paszke19pytorch}, including DP~\cite{Paszke19pytorch}, DDP~\cite{li2020pytorchddp}, and Automatic Mixed Precision (AMP)~\cite{micikevicius2018mixed}.
We also follow the linear scaling rule~\cite{Goyal@Accurate} to scale the number of parallelism in training, which is scaling the learning rate by the batch size.
\textit{Edge Device Settings:}
For hardware, we deploy a cloud-device inference system demo based on our cloud server and a resource-constrained Raspberry Pi 4B with a 4-Core Broadcom BCM2711 ARM Cortex-A72\@1.50GHz CPU and 8GB RAM, and use the USB-C power meter, POWER-Z KM003C, to record the power and energy changes on Raspberry Pi 4B during inference, as shown in Fig.~\ref{fig:inference_system}.

For on-device libraries, we choose two inference backends, i.e., PyTorch-CPU~\cite{Paszke19pytorch} and  ONNX~\cite{onnx}+ONNX Runtime~\cite{onnxruntime}, for deployment on Raspberry Pi. 
The versions are ONNX 1.14.1 and ONNX Runtime 1.15.1.

\begin{table}[htb]
\centering
    \caption{\centering{Hyperparameter setting of CNNs methods.}}
    \label{ref:hypersettings}
    \small
    \resizebox{.35\textwidth}{!}{%
    \begin{tabular}{ll}  \toprule
    \multirow{1}*{Method}
    &\makecell[c]{Reduction, Coefficient \& Range}
    \\ \midrule
    {MonoCNN}&$\bm{e}=\{2,4,8,16\}, \eta\in[1,7]$\\ 
    GhostNet& $r=2, d=3$\\  
    SineFM& $r=2, t=5, \omega\in[1,2], \phi\in[1,5]$\\ 
    \midrule
    GroupNL& $r=2, g=4, \omega\in[1,2], \phi\in[1,5]$\\ 
    \bottomrule
    \end{tabular}
}
\end{table}

\subsection{Experimental Results}
\begin{table}[htb]
    \centering
    \caption{\centering{Performance Comparison on CIFAR-10.}}
    \label{tab:standardresults}
    \small
    \resizebox{.44\textwidth}{!}{%
    \begin{tabular}{llccc}  \toprule
        &Method
        & \makecell[c]{\#Params (M)} 
        & \makecell[c]{\#FLOPs (M)}
        & \makecell[c]{Top-1 Acc (\%)}
        \\ \midrule
        \multirow{5}*{\rotatebox{90}{\makecell[c]{VGG11}}}&
        Vanilla CNN & 9.23 & 152.92 & 91.89\\  
        &MonoCNN& 0.66 & 162.18 & 88.84\\ 
        &GhostNet& 4.64 & 77.22 & \bf{92.65}\\   
        &SineFM& 4.64 & 77.68 & 91.27\\
        &\cellcolor{gray!20}GroupNL& \cellcolor{gray!20}4.62 & \cellcolor{gray!20}\bf{76.62} &         \cellcolor{gray!20}91.62 \\ 
        \midrule
        \multirow{5}*{\rotatebox{90}{\makecell[c]{ResNet-18}}}&
        Vanilla CNN & 11.17 & 556.04 & 95.04\\  
        &MonoCNN& 2.58 & 565.51 & 92.05\\ 
        &GhostNet& 5.61 & 281.13 & 93.36\\  
        &SineFM& 5.57 & 280.81 & 94.36\\
        &\cellcolor{gray!20}GroupNL& \cellcolor{gray!20}5.60 & \cellcolor{gray!20}\bf{279.49} & \cellcolor{gray!20}\bf{95.12}\\ 
        \midrule
        \multirow{5}*{\rotatebox{90}{\makecell[c]{ResNet-34}}}&
        Vanilla CNN & 21.28& 1160.43 & 95.24\\  
        &MonoCNN& 3.66 & 1180.05 & 92.26\\ 
        &GhostNet& 10.68 & 584.97 & 93.90\\   
        &SineFM& 10.65 & 586.28 & 94.77\\
        &\cellcolor{gray!20}GroupNL& \cellcolor{gray!20}10.65 & \cellcolor{gray!20}\bf{582.09} & \cellcolor{gray!20}\bf{95.45}\\ 
        \midrule
        \multirow{5}*{\rotatebox{90}{\makecell[c]{ResNet-101}}}&
        Vanilla CNN & 42.51 & 2515.09 & 95.48\\  
        &MonoCNN& 16.33 & 2544.66 & 94.97\\ 
        &GhostNet& 30.51 & 1838.52 & 95.04\\   
        &SineFM& 29.48 & 1796.49 & 95.06\\
        &\cellcolor{gray!20}GroupNL& \cellcolor{gray!20}18.46 & \cellcolor{gray!20}\bf{1158.68} & \cellcolor{gray!20}\bf{95.51}\\ 
        \bottomrule
    \end{tabular}
    }
\end{table}

\begin{table}[ht]
\centering
\caption{Performance Comparison on ImageNet-1K.\label{tab:imagenet1k}}
\resizebox{.48\textwidth}{!}{%
\begin{tabular}{@{\hspace{2mm}}ccccc@{\hspace{2mm}}}
\toprule
\multirow{2}*{\makecell[c]{Model}}
&\multirow{2}*{\makecell[c]{\#Params\\(M)}}
&\multirow{2}*{\makecell[c]{\#FLOPs\\(M)}}
&\multicolumn{2}{c}{Accuracy (\%)}\\ \cmidrule{4-5}
&&& \multirow{1}*{\makecell[c]{Top-1}} 
& \multirow{1}*{\makecell[c]{Top-5}} \\
\midrule
MobileNet-V2& 3.50 & 307.45 & 70.0 & 89.4 \\ 
MobileNet-V2-GroupNL& 3.51 & 308.60 & 69.8 & 89.3 \\ \midrule
MobileNet-V3& 5.48 & 221.59 & 73.4 & 91.6 \\
MobileNet-V3-GroupNL& 5.48 & 222.28 & 72.6 & 90.9 \\ \midrule
EfficientNet-Lite0& 4.65 & 391.93 & 72.8 & 91.4 \\
EfficientNet-Lite0-GroupNL& 4.65 & 393.08 & 72.5 & 91.1 \\ \midrule
EfficientNet-ES& 5.44 & 1797.46 & 75.9 & 93.2 \\
EfficientNet-ES-GroupNL& 5.44 & 1798.28 & 75.9 & 93.3 \\ \midrule
EfficientNet-V2-S& 21.46 & 2865.40 & 83.0 & 96.2  \\
EfficientNet-V2-S-GroupNL& 21.46 & 2867.05 & 83.1 & 96.3  \\ 
\bottomrule
\end{tabular}%
}
\end{table}

This section evaluates the robustness of the \ourmethod{}-based CNN on three types of datasets, including standard datasets (i.e., CIFAR-10 and ImageNet-1K), the corrupted dataset (i.e., CIFAR-10-C and ImageNet-C), and data under different styles (i.e., Icons50). 
\subsubsection{{Evaluation on Standard Datasets}}
Table~\ref{tab:standardresults} and Table~\ref{tab:imagenet1k} show the comparison of image classification results of different CNN models on CIFAR-10 and ImageNet-1K datasets under different methods.
We observe that \ourmethod{}-based CNNs generally outperform the accuracy of alternatives to standard CNN models.
As shown, when using ResNet-34 as the backbone, on the CIFAR-10 dataset, \ourmethod{}'s accuracy is 0.68\% higher than SineFM's, 1.55\% higher than GhostNet's, and 3.19\% higher than MonoCNN's.
When using VGG11 as the backbone, on the CIFAR-10 dataset, \ourmethod{}'s accuracy is 0.35\% higher than SineFM's.
The main reason is that \ourmethod{} generates diversified feature maps using the NLFs with different hyperparameter configurations.
In contrast, SineFM generates feature maps using a single NLF and is unable to generate diverse feature maps.
GhostNet first generates some feature maps using standard Conv operations and then uses cheap linear operations to generate other feature maps based on these generated feature maps, which limits the diversity of the feature maps.
Benefiting from its ability to generate diverse feature maps, \ourmethod{} achieves the highest accuracy among many alternatives.

To maintain the number of model parameters and FLOPs, the lightweight CNN (e.g., MobileNet, EfficientNet-ES) replaces all depthwise Convs with sparse \ourmethod{} models.
As shown in Table~\ref{tab:imagenet1k}, Sparse \ourmethod{}-based CNNs achieve similar accuracy to standard lightweight CNN models on ImageNet-1K.
As shown, \ourmethod{}'s accuracy is 0.2\% lower than MobileNet-V2's and achieves the same accuracy as EfficientNet-ES.
This effectively validates that sparse \ourmethod{} can serve as a lightweight alternative to depthwise Conv.

\subsubsection{{Evaluation on Corrupted Datasets}}
\begin{table}[t]
    \centering
    \caption{Robustness on CIFAR-10-C. 
    }
    \label{table:CIFAR-10-C}
        \scalebox{.9}{
            \begin{tabular}{llccccc} \toprule
                &Method&  Noise& Blur& Weather&Digital &mAcc ($\uparrow$) 
                \\ \midrule
                \multirow{5}*{\rotatebox{90}{\makecell[l]{ResNet-18}}}&Vanilla CNN& $52.60$& $72.78$& $84.55$& $80.70$&$ 72.58$ \\ 
                &MonoCNN& $56.36$& $63.24$& $77.79$& $75.17$&$68.14$  \\ 
                &GhostNet& $59.26$& $72.57$& $82.99$& $79.95$&$73.69$ \\ 
                &SineFM& $53.77$& $68.60$& $84.63$& $80.94$&$71.99$  \\ 
                &\cellcolor{gray!20}GroupNL& \cellcolor{gray!20}$57.55$& \cellcolor{gray!20}$72.96$& \cellcolor{gray!20}$85.25$& \cellcolor{gray!20}$80.68$&\cellcolor{gray!20}\bm{$74.11$} \\ 
                \midrule
                \multirow{5}*{\rotatebox{90}{\makecell[l]{ResNet-34}}}&Vanilla CNN& $53.72$& $71.70$& $85.08$& $82.27$&$73.19$\\
                &MonoCNN& $57.16$& $63.50$& $78.78$& $75.81$&$68.81$\\ 
                &GhostNet& $60.30$& $74.57$& $84.10$& $80.71$&$74.92$\\ 
                &SineFM& $59.20$& $71.72$& $86.11$& $81.83$&$74.72$\\ 
                &\cellcolor{gray!20}GroupNL& \cellcolor{gray!20}$59.84$& \cellcolor{gray!20}$73.64$& \cellcolor{gray!20}$86.01$& \cellcolor{gray!20}$81.26$&\cellcolor{gray!20}\bm{$75.19$}\\ 
                 \midrule
                 \multirow{5}*{\rotatebox{90}{\makecell[l]{ResNet-101}}}&Vanilla CNN& $53.97$& $73.05$& $86.02$& $81.90$&$74.78$\\ 
                 &MonoCNN& $49.03$& $74.36$& $86.60$& $82.00$&$74.26$\\ 
                 &GhostNet& $55.03$& $74.00$& $86.13$& $82.45$&$74.40$\\ 
                 &SineFM& $58.33$& $72.08$& $85.08$& $81.44$&$74.23$\\ 
                 &\cellcolor{gray!20}GroupNL& \cellcolor{gray!20}$57.90$& \cellcolor{gray!20}$71.22$& \cellcolor{gray!20}$86.45$& \cellcolor{gray!20}$81.17$&\cellcolor{gray!20}\bm{$75.04$} \\ 
                \midrule
                \multirow{5}*{\rotatebox{90}{\makecell[l]{VGG11}}}&Vanilla CNN&$65.37$& $75.45$& $82.06$& $80.48$&$75.84$\\ 
                &MonoCNN&$42.74$& $68.73$& $78.08$& $76.58$&$66.53$\\ 
                &GhostNet&$52.91$& $70.28$&$80.05$& $78.65$&$70.47$\\ 
                &SineFM&$62.31$& $72.46$& $82.46$& $82.61$&$74.96$\\ 
                &\cellcolor{gray!20}GroupNL&\cellcolor{gray!20}$64.10$& \cellcolor{gray!20}$77.24$& \cellcolor{gray!20}$83.46$& \cellcolor{gray!20}$83.87$&\cellcolor{gray!20}\bm{$77.17$} \\ 
                \bottomrule
            \end{tabular}
        }
\end{table}
%

\begin{table*}[ht]
\centering
\caption{Robustness Performance Comparison on ImageNet-C.\label{tab:imagenetc}}
\resizebox{1.\textwidth}{!}{%
\begin{tabular}{@{\hspace{2mm}}ccccccccccccccccccccc@{\hspace{2mm}}}
\toprule
\multirow{3}*{\makecell[c]{Model}}
&\multirow{3}*{\makecell[c]{mCE\\($\downarrow$)}}
&\multicolumn{18}{c}{Corruption Error ($\downarrow$)}\\ \cmidrule{3-21}
&& \multirow{2}*{\makecell[c]{Gauss.\\Noise}} 
& \multirow{2}*{\makecell[c]{Shot\\Noise}} 
& \multirow{2}*{\makecell[c]{Impulse\\Noise}} 
& \multirow{2}*{\makecell[c]{Defocus\\Blur}} 
& \multirow{2}*{\makecell[c]{Glass\\Blur}} 
& \multirow{2}*{\makecell[c]{Motion\\Blur}} 
& \multirow{2}*{\makecell[c]{Zoom\\Blur}} 
& \multirow{2}*{\makecell[c]{Snow}} 
& \multirow{2}*{\makecell[c]{Frost}} 
& \multirow{2}*{\makecell[c]{Fog}} 
& \multirow{2}*{\makecell[c]{Bright.}} 
& \multirow{2}*{\makecell[c]{Contrast}} 
& \multirow{2}*{\makecell[c]{Elastic\\Trans.}} 
& \multirow{2}*{\makecell[c]{Pixelate}} 
& \multirow{2}*{\makecell[c]{JPEG\\Comp.}}
& \multirow{2}*{\makecell[c]{Speckle\\Noise}}
& \multirow{2}*{\makecell[c]{Gauss.\\Blur}}
& \multirow{2}*{\makecell[c]{Spatter}}
& \multirow{2}*{\makecell[c]{Saturate}}
\\
\\ \midrule
Mobi-V2&  64.109 & 73.044 & 73.420 & 73.202 & 72.645 & 78.768 & 66.340 & 71.765 & 70.056 & 65.514 & 58.210 & 37.570 & 61.305 & 61.230 & 76.440 & 52.273 & 64.376 & 68.692 & 50.543 & 42.685
\\ 
Mobi-V2-G& \textbf{63.474} & 73.338 & 74.317 & 75.144 & \textcolor{blue}{70.535} & \textcolor{blue}{77.537} & \textcolor{blue}{65.395} & \textcolor{blue}{71.093} & \textcolor{blue}{69.732} & \textcolor{blue}{65.509} & 59.516 & 37.782 & \textcolor{blue}{60.291} & \textcolor{blue}{59.023} & \textcolor{blue}{66.907} & \textcolor{blue}{51.851} & 65.288 & \textcolor{blue}{66.930} & 52.895 & 42.928
\\ \midrule
Mobi-V3& 58.501 & 67.779 & 68.460 & 66.505 & 66.762 & 74.732 & 58.622 & 65.558 & 62.274 & 60.467 & 54.441 & 33.461 & 58.942 & 56.282 & 70.918 & 44.757 & 58.332 & 62.887 & 44.492 & 37.376
\\
Mobi-V3-G& \textbf{58.263} & \textcolor{blue}{66.152} & \textcolor{blue}{67.532} & 67.459 & 68.143 & \textcolor{blue}{73.531} & \textcolor{blue}{58.612} & 65.985 & 62.770 & \textcolor{blue}{60.038} & \textcolor{blue}{53.293} & 33.786 & \textcolor{blue}{58.410} & \textcolor{blue}{56.110} & \textcolor{blue}{61.212} & 48.098 & 58.707 & 64.125 & 44.745 & 38.278
\\ \midrule
Eff-Lite0 & \textbf{60.141} & 70.855 & 71.454 & 66.648 & 68.248 & 75.986 & 63.609 & 68.626 & 64.140 & 61.470 & 53.814 & 34.310 & 56.193 & 59.814 & 70.032 & 46.916 & 61.542 & 64.089 & 46.192 & 38.750
\\
Eff-Lite0-G& 61.034 & \textcolor{blue}{67.958} & \textcolor{blue}{69.381}& 68.320 & \textcolor{blue}{67.141} & 77.247 & \textcolor{blue}{62.924} & 70.036 & 65.105 & 61.960 & 54.508 & 34.728 & 66.014 &  \textcolor{blue}{59.419} & 74.681 & 48.368 & \textcolor{blue}{60.775} & \textcolor{blue}{63.917} & 48.138 & 39.002
\\ \midrule
Eff-ES& 55.033 & 62.052 & 63.339 & 65.818 & 64.430 & 73.497 & 58.716 & 65.290 & 59.704 & 54.999 & 47.713 & 30.732 & 45.922 & 57.111 & 60.785 & 44.755 & 54.082 & 60.322 & 41.914 & 34.451
\\
Eff-ES-G& \textbf{53.958} & \textcolor{blue}{60.843} & \textcolor{blue}{61.825} & \textcolor{blue}{65.581} & \textcolor{blue}{62.466} & \textcolor{blue}{72.968} & \textcolor{blue}{58.113} & \textcolor{blue}{63.870} & \textcolor{blue}{57.496} & \textcolor{blue}{54.063} & \textcolor{blue}{47.302} & \textcolor{blue}{30.174} & 46.991 & \textcolor{blue}{55.622} & \textcolor{blue}{58.412} & \textcolor{blue}{43.345} & \textcolor{blue}{53.372} & \textcolor{blue}{58.514} & \textcolor{blue}{40.574} & \textcolor{blue}{33.670}
\\ \midrule
Eff-V2-S& \textbf{41.551} & 41.165 & 42.186 & 40.098 & 51.918 & 63.25 & 43.122 & 53.604 & 43.947 & 41.364 &42.209 & 24.209 & 35.391 & 47.551 & 42.902 & 34.116 & 36.144 & 48.70 & 30.879 & 26.714   
\\
Eff-V2-S-G& 41.720 & 42.267 & 43.691 & 42.394 & \textcolor{blue}{51.855} & \textcolor{blue}{63.165} & \textcolor{blue}{43.023} & \textcolor{blue}{52.579} & 45.158 & 42.225 & \textcolor{blue}{39.69} & \textcolor{blue}{24.017} &   37.828 & 47.602 & \textcolor{blue}{37.545} & \textcolor{blue}{33.616} & 37.396 & 49.422 & 32.361 & 26.842
\\ 
\bottomrule
\end{tabular}%
}
\end{table*}
Table~\ref{table:CIFAR-10-C} and Table~\ref{tab:imagenetc} show the performance comparison of different models for each method on the corrupted data. We observe that the \ourmethod{}-based CNN models achieve the highest accuracy among alternatives to standard Conv.
As shown in Table~\ref{table:CIFAR-10-C}, when ResNet-34 is used as the backbone, the accuracy of GroupNL is 0.47\% higher than that of SineFM, 6.38\% higher than that of MonoCNN, 0.27\% higher than that of GhostNet, and even 2\% higher than the accuracy of standard ResNet-34.
When VGG11 is used as the backbone, the accuracy of GroupNL is 2.21\% higher than that of SineFM, 10.64\% higher than that of MonoCNN, 6.7\% higher than that of GhostNet, and 1.33\% higher than the accuracy of standard VGG11.
There are two reasons for this: 
(i) The rules of the NLF can regularize the model and improve the generalization ability of the model.
(ii) NLFs equipped with different hyperparameters can generate diverse feature maps and improve the knowledge expression ability of the model.
Therefore, GroupNL achieves the highest accuracy.

Sparse \ourmethod{} also achieves considerable performance as a replacement for lightweight CNN models on ImageNet-C.
As shown in Table~\ref{tab:imagenetc}, the models based on sparse \ourmethod{} even achieve higher performance compared to the standard MobileNet-V2/V3, and EfficientNet-ES.
This further validates that sparse \ourmethod{} can serve as a lightweight alternative to depthwise Convs when dealing with corrupted datasets.
\begin{table}[t]
    \centering
    \caption{Robustness to Different Styles on Icons-50. 
    }
    \label{table:Icons50}
        \scalebox{.85}{
            \begin{tabular}{llccccc} \toprule
                &Method&Apple& Facebook& Google&Samsung&mAcc ($\uparrow$)
                \\ \midrule
                \multirow{5}*{\rotatebox{90}{\makecell[l]{ResNet-18}}}&Vanilla CNN&$93.66$& $87.64$& $83.20$& $80.28$&$86.19$\\ 
                &MonoCNN&$93.02$& $85.37$&$81.60$& $78.59$&$84.65$\\ 
                &GhostNet&$92.07$& $85.09$& $78.98$& $77.61$&$83.43$\\ 
                &SineFM&$94.84$& $88.72$& $84.96$& $82.27$&$87.70$\\
                &\cellcolor{gray!20}GroupNL&\cellcolor{gray!20}$94.52$&\cellcolor{gray!20}$91.05$ &\cellcolor{gray!20}$86.27$ &\cellcolor{gray!20}$84.36$&\cellcolor{gray!20}\bm{$89.05$}\\
                \midrule
                \multirow{5}*{\rotatebox{90}{\makecell[l]{ResNet-34}}}&Vanilla CNN&$92.77$& $86.26$& $83.81$& $82.44$&$86.32$\\ 
                &MonoCNN&$91.46$& $85.88$& $81.55$& $79.23$&$84.53$\\
                &GhostNet&$89.87$& $84.70$&$79.61$& $73.76$&$81.98$\\
                &SineFM&$94.74$& $88.72$& $84.88$& $81.46$&$87.45$\\%
                &\cellcolor{gray!20}GroupNL&\cellcolor{gray!20}$95.19$ &\cellcolor{gray!20}$90.98$ &\cellcolor{gray!20}$85.93$ &\cellcolor{gray!20}$84.36$&\cellcolor{gray!20}\bm{$89.12$}\\
                \midrule
                \multirow{5}*{\rotatebox{90}{\makecell[l]{ResNet-101}}}&Vanilla CNN&$93.64$& $87.70$& $83.64$& $83.13$&$87.03$\\ 
                &MonoCNN&$94.84$& $87.64$& $83.62$& $82.71$&$87.20$\\ 
                &GhostNet&$93.50$& $88.53$&$83.62$& $84.50$&$87.54$\\
                &SineFM&$90.25$& $83.65$& $79.69$& $74.67$&$82.06$\\ 
                &\cellcolor{gray!20}GroupNL&\cellcolor{gray!20}$93.98$&\cellcolor{gray!20}$90.15$ &\cellcolor{gray!20}$85.51$&\cellcolor{gray!20}$83.89$&\cellcolor{gray!20}\bm{$88.38$}\\ 
                \midrule
                \multirow{5}*{\rotatebox{90}{\makecell[l]{VGG11}}}&Vanilla CNN&$92.54$& $86.68$& $83.67$& $80.34$&$85.81$\\ 
                &MonoCNN&$89.55$& $81.10$& $79.82$& $75.41$&$81.47$\\ 
                &GhostNet&$91.40$& $88.15$&$80.71$& $79.33$&$84.90$\\ 
                &SineFM&$94.74$& $89.87$& $85.59$& $82.17$&$88.09$\\ 
                &\cellcolor{gray!20}GroupNL&\cellcolor{gray!20}$94.23$ &\cellcolor{gray!20}$91.05$ &\cellcolor{gray!20}$85.30$ &\cellcolor{gray!20}$82.13$ &\cellcolor{gray!20}\bm{$88.18$}\\ 
                \bottomrule
            \end{tabular}
        }
\end{table}

\subsubsection{{Evaluation on Dataset under Different Styles}}
We obtain conclusions similar to those for the corrupted data; the \ourmethod{}-based CNN models achieve the highest accuracy under all backbones.
As shown in Table~\ref{table:Icons50}, when ResNet-18 is used as the backbone, the accuracy of \ourmethod{} is 1.35\% higher than that of SineFM, 4.4\% higher than that of MonoCNN, 5.62\% higher than that of GhostNet, and even 2.86\% higher than the accuracy of standard ResNet-18.
when ResNet-101 is used as the backbone, the accuracy of \ourmethod{} is 6.3\% higher than that of SineFM, 1.18\% higher than that of MonoCNN, 0.84\% higher than that of GhostNet, and even 1.35\% higher than the accuracy of standard ResNet-18.
The reason for this is the same as that of our corrupted data; that is, 
(i) the nonlinear transformation function in \ourmethod{} regularizes the model; (ii) \ourmethod{} can generate diverse feature maps.
Therefore, \ourmethod{} has the highest accuracy.

\subsubsection{Training Acceleration Evaluation} 
\begin{table}
    \centering
    \caption{\centering{Comparison of CIFAR-10 Data Parallel Training Time in Different CNN methods in RTX 2080Ti GPUs.}}
    \label{Table:trainingAccelerate_all_cnn}
    \resizebox{.45\textwidth}{!}{%
    \begin{tabular}{clclccc} \toprule
    \makecell[c]{No. of\\GPUs} &
    \multirow{1}*{\makecell[c]{LR}} &
    \multirow{1}*{\makecell[c]{BS}} &
    \multirow{1}*{Method}& 
    \makecell[c]{Top-1\\Acc (\%)} & 
    \makecell[c]{Training\\Time (h)}& 
    \makecell[c]{Accel.\\Ratio}\\\midrule
    \multirow{5}*{\makecell[c]{2-GPUs}} & 
    \multirow{5}*{\makecell[c]{0.1}} & 
    \multirow{5}*{\makecell[c]{64$\cdot$2}}& Vanilla CNN
    & 95.48& 9.3h&-\\ 
    & & & MonoCNN & 94.97 & 9.2h &{+0\%}\\ 
    & & & GhostNet& 95.04 & 12.7h&{-37\%}\\ 
    & & & SineFM& 95.06 & 15.4h&{-66\%}\\ 
    & & & \cellcolor{gray!20}\bf{GroupNL}& \cellcolor{gray!20}{\bf{95.51}} & \cellcolor{gray!20}\bf{8.3h}&\cellcolor{gray!20}{\bf{+11\%}}\\ 
    \bottomrule
    \end{tabular}
    }
\end{table}

We verify the training acceleration of each CNN method based on ResNet-101.
We observe that:

(i) Compared to vanilla ResNet-101, \ourmethod{} ResNet-101 achieves 11\% speedup while achieving the higher accuracy, as shown in Table~\ref{Table:trainingAccelerate_all_cnn}.
Moreover, since only two GPUs are used, the \ourmethod{} mainly accelerates model training by reducing model calculation time.

(ii) Although the GhostNet and SineFM have fewer parameters and computational complexity than vanilla CNN, the training time is longer, as shown in Table~\ref{Table:trainingAccelerate_all_cnn}.
As we analyzed in Section~\ref{ref: cnn}, although cheap Conv and BN require less computation, they are slow to compute and optimize.
Incorporating these operations into the model will significantly slow down the model's training speed.

We further compare the speed of \ourmethod{} and vanilla CNN with different numbers of GPUs under DP and DDP.

\begin{figure}
\centering
\includegraphics[width=.9\linewidth]{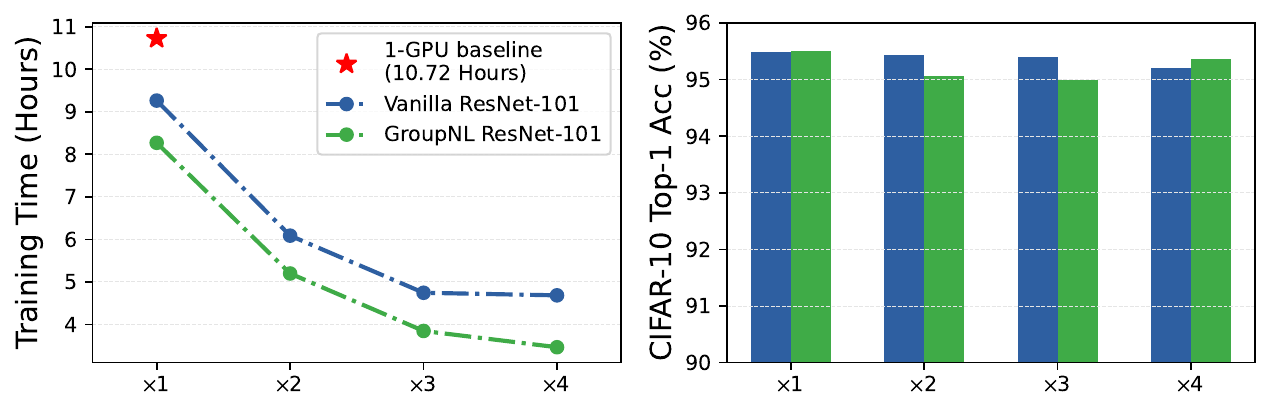}
\caption {CIFAR-10 Training Acceleration Evaluation with Data Parallelism in RTX 2080Ti GPUs.}
\label{fig:dp_accel}
\end{figure}

\emph{Training Evaluation in DP}. 
For the multi-GPUs evaluation using DP, Table~\ref{Table:trainingAccelerate_dp} and Fig.~\ref{fig:dp_accel} show the experimental settings and training results, respectively. We observe that:
\begin{table}
    \centering
    \caption{\centering{Training Settings in DP/DDP Evaluation.}}
    \label{Table:trainingAccelerate_dp}
    \resizebox{.42\textwidth}{!}{%
    \begin{tabular}{lccclc} \toprule
    &
    \multirow{1}*{\makecell[c]{Platform}} &
    \multirow{1}*{\makecell[c]{Parallelism}} &
    \multirow{1}*{\makecell[c]{No. of GPUs}} &
    \multirow{1}*{\makecell[c]{LR}} &
    \multirow{1}*{\makecell[c]{BS}}\\ 
    \midrule
    \multirow{5}*{\rotatebox{90}{\makecell[l]{DP}}}&\multirow{5}*{\makecell[c]{RTX\\2080Ti}} &
    baseline &1-GPU& 0.1 & 128$\cdot$1 \\
    &&$\times$1 &2-GPUs& 0.1 & 64$\cdot$2 \\
    &&$\times$2 &2-GPUs& 0.2 & 128$\cdot$2 \\
    &&$\times$3 &3-GPUs& 0.3 & 128$\cdot$3 \\
    &&$\times$4 &4-GPUs& 0.4 & 128$\cdot$4 \\
    \midrule
    \multirow{5}*{\rotatebox{90}{\makecell[l]{DDP}}}&\multirow{4}*{\makecell[c]{RTX\\2080Ti}}
    &baseline &1-GPU& 0.1 & 128$\cdot$1 \\
    &&$\times$2 &2-GPUs& 0.2 & 128$\cdot$2 \\
    &&$\times$4 &2-GPUs& 0.4 & 256$\cdot$2 \\
    &&$\times$8 &4-GPUs& 0.8 & 256$\cdot$4 \\
    \cmidrule{2-6}
    &\multirow{1}*{\makecell[c]{RTX4090}}
    &$\times$8 &8-GPUs& 0.001 & 128$\cdot$8 \\
    \bottomrule
    \end{tabular}
    }
\end{table}

With more GPUs, the \ourmethod{} has a larger training speed-up ratio.
As shown in Fig.~\ref{fig:dp_accel}, when the number of GPUs is 4, \ourmethod{} has the largest speed-up ratio compared to the vanilla CNN.
This is because when the number of GPUs $N$ is large in DP, gradient communication $(N\!-\!1)\mathcal{G}$ for the main GPU and $\mathcal{G}$ for the rest $(N\!-\!1)$ GPUs will increase when training a vanilla CNN.
In contrast, the \ourmethod{} with fewer gradients is less affected by the scale of GPUs.
Therefore, when a large number of GPUs are used, \ourmethod{} has a higher training speedup ratio.

\begin{figure}
\centering
\includegraphics[width=.9\linewidth]{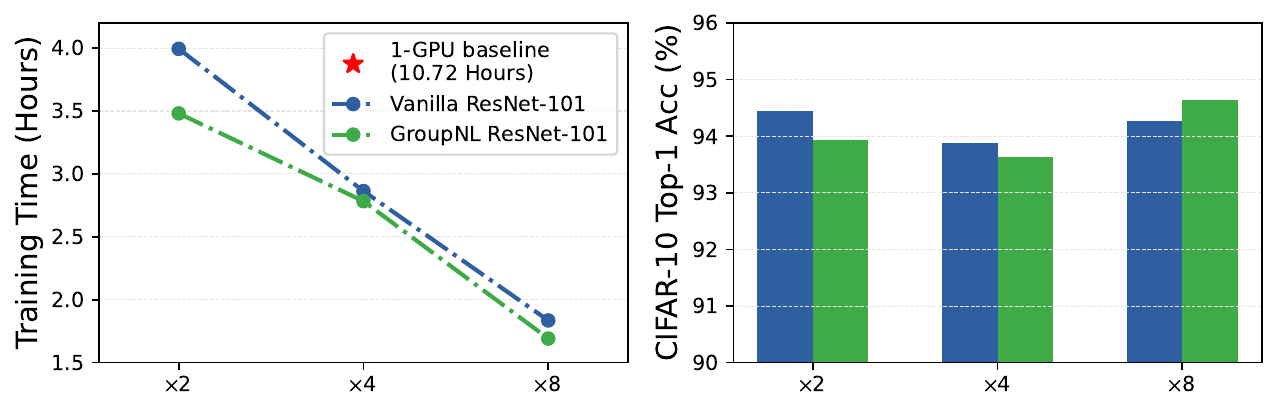}
\caption {CIFAR-10 Training Acceleration Evaluation with DDP and AMP in RTX 2080Ti GPUs.}
\label{fig:ddp_accel}
\end{figure}

\emph{Training Evaluation in DDP}.
For the multi-GPUs evaluation with DDP, Table~\ref{Table:trainingAccelerate_dp} and Fig.~\ref{fig:ddp_accel} show the experimental settings and evaluated results of DDP training with AMP, respectively. We observe that:
The DDP uses Ring-AllReduce to reduce gradient communication to $\frac{2(N-1)}{N}\mathcal{G}$ between GPUs. The \ourmethod{} also reduces training time compared to vanilla CNNs, mainly due to fewer gradients.

\begin{figure}
\centering
\includegraphics[width=.9\linewidth]{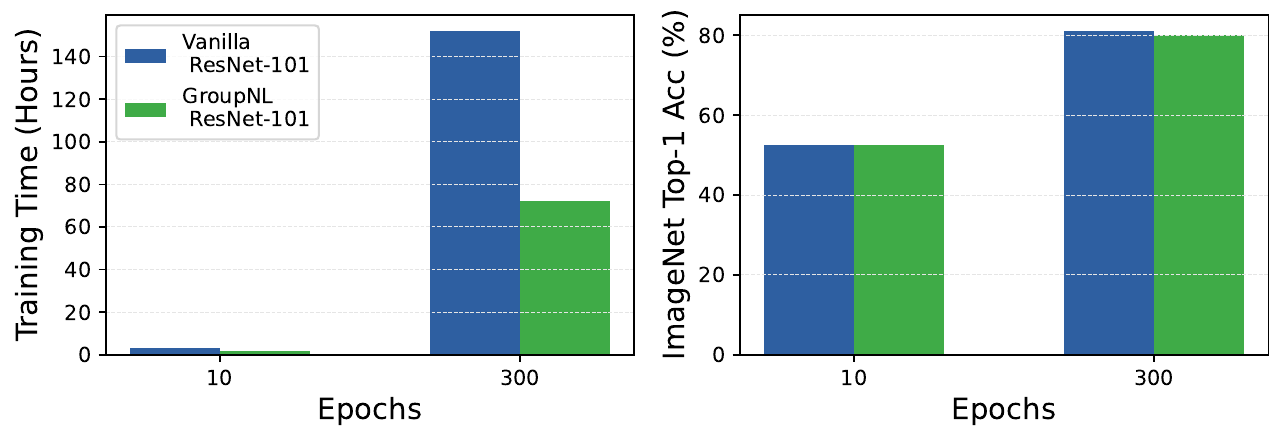}
\caption {ImageNet-1K Training Acceleration Evaluation with DDP (Parallelism ×8) in RTX 4090 GPUs.}
\label{fig:ddp_accel_imagenet}
\end{figure}

We also show the training speedup results for ImageNet-1K training of 8 GPUs.
The \ourmethod{} reduces training time by about 53\% while achieving similar accuracy to the vanilla CNN.
As shown in Fig~\ref{fig:ddp_accel_imagenet}.
The $\times$8 DDP training vanilla CNN for 10 epochs is 3.32 hours with a Top-1 accuracy of 52.60\%, and \ourmethod{} is 1.87 hours with a Top-1 accuracy of 52.65\%.
For the $\times$8 DDP training vanilla CNN for 300 epochs is 151.93 hours with a Top-1 accuracy of 81.04\%, and \ourmethod{} is 72.11 hours with a Top-1 accuracy of 80.16\%.

\emph{Communication Cost \& Epoch Time}. We also evaluate the communication overhead and training time for 8-GPU DDP training of ResNet-101 and the corresponding GroupNL variant. 
With sizes of 44.55M and 26.37M gradients in models, respectively, the per-GPU communication cost and total 8-GPUs' communication costs are 77.96M and 623.69M for ResNet-101, and 46.14M and 369.13M for GroupNL ResNet-101, respectively.
Using the $\mathrm{epoch}=149$ training as an example, GroupNL ResNet-101 completes this epoch training in 1045.79s, with an average batch time of 0.827s, compared to 1754.07s and 1.398s for the vanilla version. 
This represents a 40\% reduction in training time and a 41\% reduction in gradients communication, highlighting the efficiency of GroupNL in acceleration.

In general, \ourmethod{} optimizes the model acceleration in terms of gradients communication and calculation time.

\subsection{Deployment and Resource Measurements}
\subsubsection{Cloud-Device Transmission Cost} 
We follow the cloud-device deployment pipeline in~\cite{xie25reframe}: 
For PyTorch-CPU, we transmit well-trained PyTorch models;
For ONNX Runtime, we convert PyTorch models to optimized ONNX models via gradient removal and BN fusion, and then transmit the ONNX models to the Raspberry Pi.

\begin{table}[ht]
\centering
\caption{Cloud-Device Transmission Model Size.\label{tab:modelsize}}
\resizebox{.42\textwidth}{!}{%
\begin{tabular}{@{\hspace{2mm}}ccc@{\hspace{2mm}}}
\toprule
\multirow{2}*{\makecell[c]{Model}}
&\multirow{2}*{\makecell[c]{PyTorch Size\\(Trained Model)}}
&\multirow{2}*{\makecell[c]{ONNX Size}}
\\
&&\\ \midrule
MobileNet-V3& 84.2MB & 20.9MB \\
MobileNet-V3-GroupNL& 84.2MB& 21.1MB \\ \midrule
EfficientNet-ES& 62.6MB & 20.7MB  \\
EfficientNet-ES-GroupNL& 62.6MB & 20.9MB \\ \midrule
ResNet-50 & 292.9MB & 97.4MB  \\ 
ResNet-50-GroupNL  & 199.1MB & 66.3MB  \\
\midrule
ResNet-101 & 510.7MB & 169.8MB \\ 
ResNet-101-GroupNL & 302.6MB & 100.7MB  \\
\bottomrule
\end{tabular}%
}
\end{table}

As shown in Table~\ref{tab:modelsize}, GroupNL maintains nearly identical sizes for lightweight CNNs like MobileNet-V3 and EfficientNet-ES in both PyTorch and ONNX formats.
Also, it achieves substantial size reductions in larger-size CNNs, decreasing ResNet-50 by approximately 32\% and ResNet-101 by about 41\%.
This demonstrates GroupNL's advantage for efficient cloud-device transmission.

\subsubsection{Resource Evaluations in On-Device Inference} 
We infer 100 images randomly selected from ImageNet-1K validation datasets on a resource-constrained Raspberry Pi to calculate the inference time, average Frames Per Second (FPS), memory usage, average power, and energy consumption.
The average FPS computation follows the setting of this work~\cite{xie2024multivision}.

\begin{table}[ht]
\centering
\caption{On-Device Evaluation with PyTorch-CPU.\label{tab:measurement_torchcpu}}
\resizebox{.48\textwidth}{!}{%
\begin{tabular}{@{\hspace{2mm}}ccccccc@{\hspace{2mm}}}
\toprule
\multirow{3}*{\makecell[c]{Model}}
&\multirow{3}*{\makecell[c]{Time\\(Sec)}}
&\multirow{3}*{\makecell[c]{Avg.\\FPS ($\uparrow$)}}
&\multicolumn{3}{c}{Avg. Resource Usage} 
&\multirow{3}*{\makecell[c]{Acc\\(\%)}}\\ \cmidrule{4-6}
&&& \multirow{2}*{\makecell[c]{Mem.\\(MB)}} 
& \multirow{2}*{\makecell[c]{Power\\(W)}} 
& \multirow{2}*{\makecell[c]{Energy\\(Wh)}}
\\
\\ \midrule
Mobi-V3& 52.9 & 1.89 & 180.0 & 6.144 & 0.0973 & 86 \\
Mobi-V3-G& 43.5 & 2.30 & 181.0 & 6.034 & 0.0805 & 84 \\ \midrule
Eff-ES& 122.0 & 0.82 & 187.9& 6.101 & 0.2102 & 88 \\
Eff-ES-G& 95.1 & 1.05 & 206.1 & 5.866 & 0.1629 & 88 \\ \midrule
RS-50 & 106.7 & 0.94 & 269.4 & 5.171 & 0.1623 & 91 \\ 
RS-50-G & 86.7 & 1.15 & 228.8 & 5.452 & 0.1363 & 92 \\
\midrule
RS-101 & 192.7 & 0.52 & 340.0 & 5.125 & 0.2819 & 94 \\ 
RS-101-G & 146.4 & 0.68 & 262.0 & 5.518 & 0.2345 & 94 \\
\bottomrule
\end{tabular}%
}
\end{table}

\begin{figure}
\centering
\includegraphics[width=1.\linewidth]{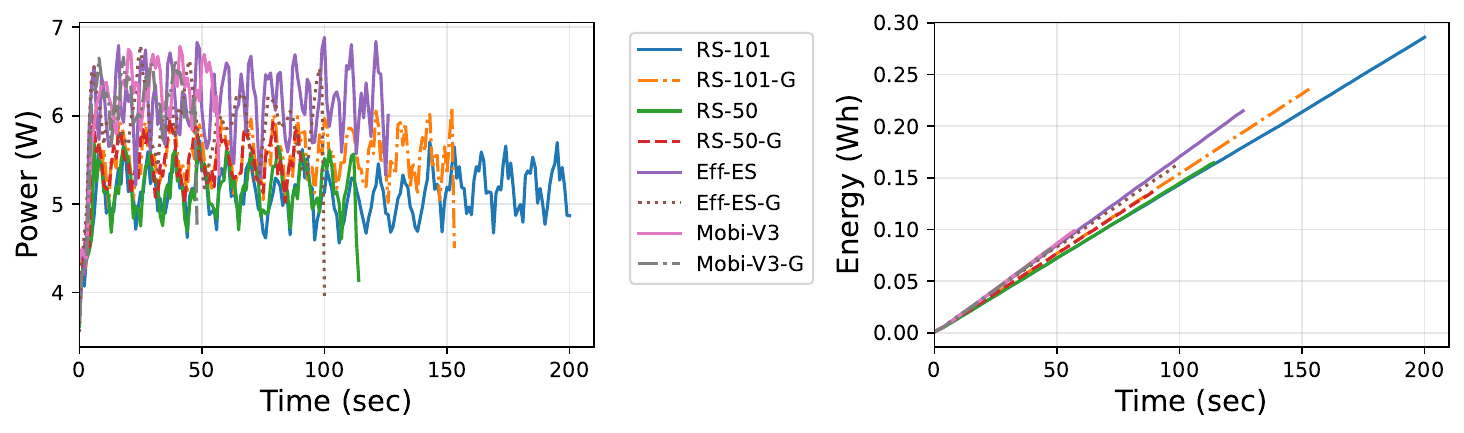}
\caption {Power and Energy Consumption in PyTorch-CPU.}
\label{fig:torchcpu_power_energy}
\end{figure}

\emph{PyTorch-CPU Backend}. As shown in Table~\ref{tab:measurement_torchcpu}, GroupNL significantly boosts inference efficiency on the Raspberry Pi. 
For example, in EfficientNet-ES, it reduces inference time and energy consumption by up to 22\% and 22.5\%, respectively. 
It also maintains competitive accuracy and reduces memory overhead in larger models like ResNet-50 and ResNet-101.
As shown in Fig.~\ref{fig:torchcpu_power_energy}, for power consumption (left), GroupNL yields shorter inference times and lower peak power than its counterparts;
For energy consumption (right), the shorter inference time of GroupNL results in lower total energy.

\begin{table}[ht]
\centering
\caption{On-Device Evaluation with ONNX Runtime.\label{tab:measurement}}
\resizebox{.48\textwidth}{!}{%
\begin{tabular}{@{\hspace{2mm}}ccccccc@{\hspace{2mm}}}
\toprule
\multirow{3}*{\makecell[c]{Model}}
&\multirow{3}*{\makecell[c]{Time\\(Sec)}}
&\multirow{3}*{\makecell[c]{Avg.\\FPS ($\uparrow$)}}
&\multicolumn{3}{c}{Avg. Resource Usage} 
&\multirow{3}*{\makecell[c]{Acc\\(\%)}}\\ \cmidrule{4-6}
&&& \multirow{2}*{\makecell[c]{Mem.\\(MB)}} 
& \multirow{2}*{\makecell[c]{Power\\(W)}} 
& \multirow{2}*{\makecell[c]{Energy\\(Wh)}}
\\
\\ \midrule
Mobi-V3& 13.0 & 7.67 & 174.0 & 6.347  &0.0277  & 86 \\ 
Mobi-V3-G& 15.8 & 6.35 & 178.0 & 6.290 & 0.0330 
  & 84 \\ \midrule
Eff-ES& 29.1 & 3.44 & 176.0 &6.698&	0.0588& 88 \\
Eff-ES-G& 32.8 & 3.04 & 183.0 & 6.671 &0.0660 & 88 \\ \midrule
RS-50 & 44.5 & 2.25 & 270.9 & 6.772 &0.0922 & 91 \\ 
RS-50-G & 35.4 & 2.83 & 236.0 & 6.795 &0.0719  & 92 \\
\midrule
RS-101 & 72.3 & 1.38 & 343.9 &6.841 &0.1497 & 94 \\ 
RS-101-G & 55.9 & 1.79 & 272.9 & 6.766 & 0.1152  & 94 \\
\bottomrule
\end{tabular}%
}
\end{table}

\begin{figure}
\centering
\includegraphics[width=1.\linewidth]{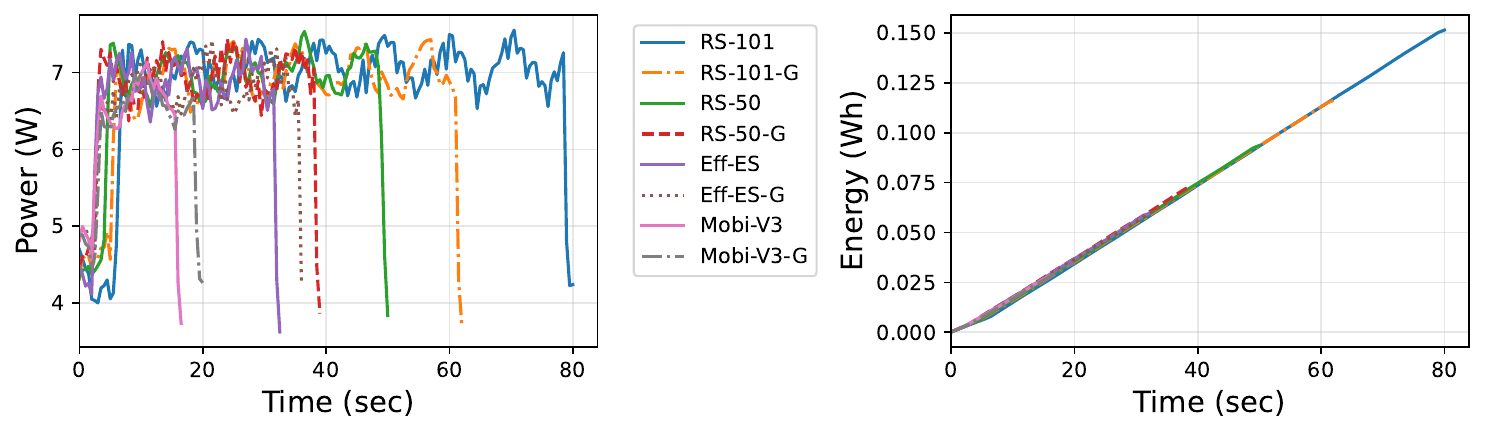}
\caption {Power and Energy Consumption in ONNX Runtime.}
\label{fig:onnxrt_power_energy}
\end{figure}

\emph{ONNX Runtime Backend}. As shown in Table~\ref{tab:measurement}, GroupNL improves inference efficiency for larger models while showing limited gains on lightweight CNNs when using ONNX Runtime.
For instance, in ResNet-101, it reduces inference time and energy consumption by approximately 22.7\% and 23\%, respectively.
However, in lightweight CNNs such as EfficientNet-ES, it currently incurs a slight increase in latency and energy use. 
This is because the converted GroupNL ONNX model does not yet support efficient tensor manipulation operators, e.g., one $\tt{torch.repeat}$ is converted into multiple $\tt{onnx.tile}$ and $\tt{onnx.expand}$ operations, introducing overhead that diminishes its advantage compared with depthwise Conv. 
A similar power and energy consumption is reported in Fig.~\ref{fig:onnxrt_power_energy}.

\subsection{Ablation Studies}

\subsubsection{Selection of Nonlinear Functions}
Choosing appropriate nonlinear transformations and selecting appropriate hyperparameters for the NLF is highly related to the performance of the GroupNL model.

{\emph{Nonlinear Functions}}: 
It is worth noting that vanilla convolution operations utilize nonlinear functions for activation, which motivates us to consider the power of nonlinear transformation in feature map generation.
Theoretically, all nonlinear functions can be used as NLFs in GroupNL to generate other feature maps based on the seed feature maps.

\begin{figure}[tbp]
	\centering
	\subfloat[\centering{Accuracy of different NLFs.}]{
        \label{Fig.3(a)}
        \includegraphics[scale = 0.26]{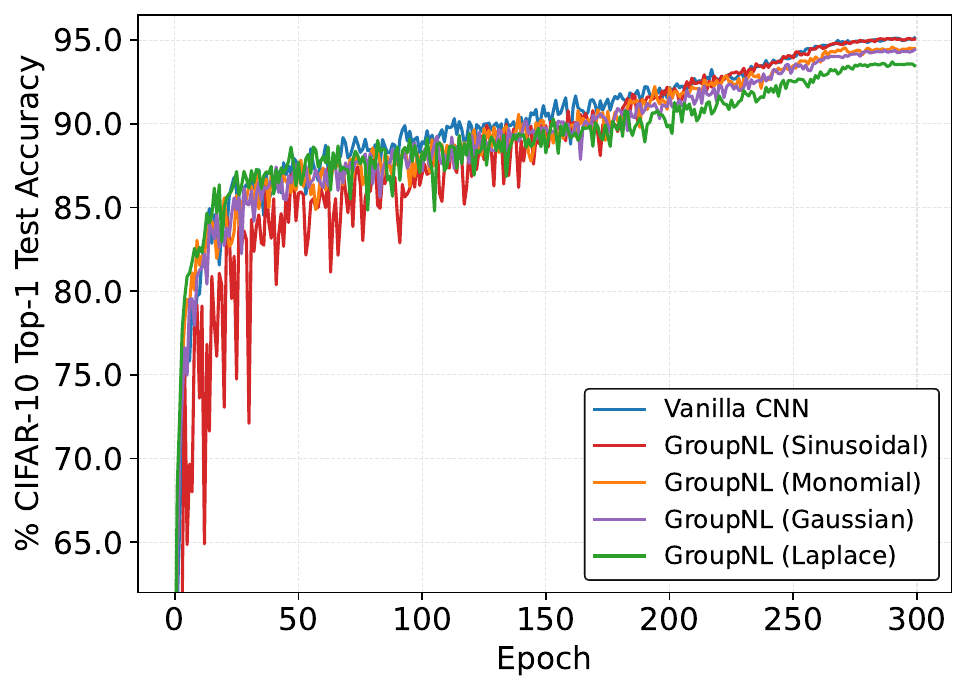}}
	\subfloat[\centering{Loss of different NLFs.}]{
        \label{Fig.3(b)}
        \includegraphics[scale = 0.26]{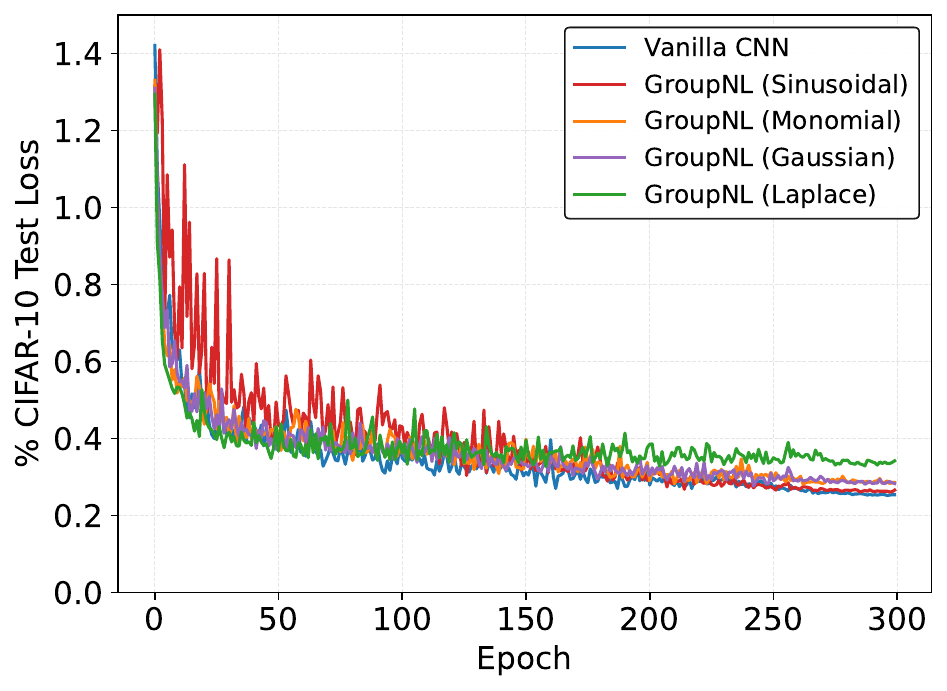}}\\	
	\caption{Comparison of different NLFs in GroupNL and vanilla ResNet-18 on the CIFAR-10 dataset.}
 \label{ref:AccuracyvsLoss}
\end{figure}

Referring to the aforementioned research about nonlinear transformations~\cite{Ding@Towards, Lu@Seed}, we select four different optimal nonlinear functions verified in these works, including the \emph{Sinusoidal} function, \emph{Monomial} function, \emph{Gaussian}, and \emph{Laplace}. 
Table~\ref{ref:tab_NLF_hyperparam} shows the specific formulas and corresponding hyperparameters of the above nonlinear functions.
Furthermore, Fig.~\ref{ref:AccuracyvsLoss} illustrates the performance results of the GroupNL models with ResNet-18 backbone in four different NLFs on the CIFAR-10 dataset.
\renewcommand{\thetable}{\arabic{table}}
\begin{table}
    \centering
    \caption{Nonlinear transformation functions.}
    \resizebox{.4\textwidth}{!}{%
    \label{ref:tab_NLF_hyperparam}
    \begin{tabular}{llc} \toprule
         \makecell[l]{Nonlinear}& Formulation&Hyperparameters\\ \midrule
         Sinusoidal& $\mathrm{sin}(\omega (x + \phi))$ &$\omega, \phi\in \mathbb{R}$\\
         Monomial& $\mathrm{sign}(x){|x|}^\eta$& $\eta\in \mathbb{R}$\\
         Gaussian& $e^{-{(\epsilon x)}^2}$& $\epsilon\in \mathbb{R}$\\
         Laplace& $\frac{\epsilon}{2} e^{-(\epsilon x)}$& $\epsilon\in \mathbb{R}$\\ \bottomrule
    \end{tabular}}    
\end{table}

In this case, we can observe that the GroupNL model with the Sinusoidal function outperforms the other three NLFs and also has a similar performance to the vanilla CNN model.
As a result, in our experiments, we use the Sinusoidal function as the NLF to generate feature maps.

We offer a conjecture: due to the periodicity of the Sinusoidal function and its different frequencies and phase hyperparameters, it can generate a richer variety of ``similar but phase-different patterns'' compared to other functions, which meets the requirement of \ourmethod{} to generate similar but diverse feature maps.
The reasons why other functions are worse may be: Monomial only performs power transformations and does not generate new spatial patterns, resulting in insufficient diversity; Gaussian function has too strong locality, with values only near the function center, and other values are suppressed to 0, which cannot guarantee the similarity of the generated images; Laplace function is similar to Gaussian.

\subsubsection{Selection of Hyperparameters and Coefficients} 

{\emph{Analysis of Sparse Seed Filters in GroupNL}}:
As described in Section~\ref{ref-groupnl-algo}, we designate that the seed filters ${\bm{W}}_\mathrm{seed}$ in basic structure (e.g., ResNet-18) utilize the standard convolution, and the seed filters in bottleneck structure (e.g., ResNet-101) are sparse with convolutional groups of $\mathrm{Gcd}(c_\mathrm{in}, c_\mathrm{seed})$.
The sparse seed filter experimental results are shown in Table~\ref{ref:sparse_and_r}.
Applying sparsity to the seeded convolution of ResNet-18 can introduce an extremely low number of parameters of 0.05M and approximately 7M FLOPs. 
However, performance is severely degraded.
As a comparison, the ResNet-101 with sparsity can maintain the performance with the lowest number of parameters and FLOPs. 
Therefore, the sparse seed filters are not used in the basic structure.

Additionally, Sparse GroupNL modules are designated to replace the depthwise convolutions in lightweight CNNs~\cite{mobilenetv2, Howard@Searching, efficientnet20lite, tan19efficientnet, efficientnet19edgetpu, tan21efficientnetv2} to maintain the compactness of these models in terms of number of parameters and FLOPs.

\begin{table}
    \centering
    \caption{Ablation Studies of w/ and w/o Sparse Seed Filters and Reduction Ratio $r$ in GroupNL.}
    \label{ref:sparse_and_r}
    \resizebox{.4\textwidth}{!}{%
    \begin{tabular}{cccccc} \toprule
        \makecell[c]{Model}& $g$&$r$ & \makecell[c]{\#Params\\(M)} & \makecell[c]{\#FLOPs\\(M)} & \makecell[c]{Top-1\\Acc (\%)}\\ \midrule
        \multirow{3}*{\makecell[c]{ResNet-18\\(w/o sparse)}} & \multirow{3}*{4} & 2 & 5.6 & 279 & 95.12 \\
         & & 4 & 2.8 & 141 &  94.33 \\
         & & 8 & 1.4 & 72 & 92.94 \\\midrule
        \multirow{3}*{\makecell[c]{ResNet-18\\(w/ sparse)}} & \multirow{3}*{4} & 2 & 0.05 & 6.9 & 89.66 \\
         & & 4 & 0.05 & 7.0 & 87.17 \\
         & & 8 & 0.05 & 7.1 & 87.67 \\\midrule
        \multirow{4}*{\makecell[c]{ResNet-101\\(w/o sparse)}} & \multirow{4}*{4} & 2 & 29.4 & 1790 & 95.57\\ 
         & & 4 & 23.9 & 1472 & 95.21 \\
         & & 8 & 21.2 & 1312 & 95.06 \\ 
         & & 16 & 19.8 & 1233 & 93.73 \\\midrule
        \multirow{4}*{\makecell[c]{ResNet-101\\(w/ sparse)}} & \multirow{4}*{4} & 2 & 18.5 & 1159& 95.51\\ 
         & & 4 & 18.5 & 1159 & 95.40  \\
         & & 8 & 18.5 & 1159&95.10 \\ 
         & & 16 & 18.5 & 1159& 93.83 \\ \bottomrule 
    \end{tabular}
    }
\end{table}

{\emph{Hyperparameters and Coefficients of GroupNL}}:
We designate that the hyperparameters of NLFs are randomly initialized and non-learnable during model training.
This trick reduces the number of gradients for optimization in training while preventing the NLFs from losing the data-agnostic properties.
The Sinusoidal function has two hyperparameters $\omega$ and $\phi$, we randomly choose them via $\tt{torch.rand}$ between $[1,2]$ and $[1,5]$, respectively.

In addition, we also explore the reduction ratio $r\!=\!c_\mathrm{out}/c_\mathrm{seed}$ and the number of splitting groups $g$.
Ablation studies of different configurations on $r$ and $g$ are shown in Table~\ref{ref:sparse_and_r} and Table~\ref{ref:gcomp}, respectively.
As shown in Table~\ref{ref:sparse_and_r}, the model's accuracy decreases as the reduction ratio increases.
As shown in Table~\ref{ref:gcomp}, with the fixing $r$, the change in $g$ results in model performance improvement in proper $g$.

Besides, for the ResNet-101, the number of parameters and FLOPs are very similar for different $r$ settings.
This is due to the bottleneck structure in ResNet-101 using the sparse seed filters with convolutional groups of $\xi=\mathrm{Gcd}(c_\mathrm{in}, c_\mathrm{seed})$.
For example, assuming a $k^2$ convolution with same input and output channels $c_\mathrm{in}\!=\!c_\mathrm{out}\!=\!512$ and feature maps size $wh$ in the bottleneck. 

In the case of $r\!=\!2$, the number of parameters is $\mathcal{P}_{r2} \!=\! c_\mathrm{in}c_\mathrm{seed}k^2\frac{1}{ \xi}\!=\!512\cdot256k^2\frac{1}{256}\!=\!512k^2$, and the FLOPs is $\mathcal{F}_{r2} \!=\! whc_\mathrm{in}c_\mathrm{seed}k^2\frac{1}{ \xi}\!=\!wh\cdot512\cdot256k^2\frac{1}{256}\!=\!512k^2wh$.
In the cases of $r\!=\!\{4,16,32,\cdots,256,512\}$, the results are also $\mathcal{P}_{r}\!=\!512k^2$ and $\mathcal{F}_{r}\!=\!512k^2wh$ because of the fraction $c_\mathrm{seed}$ are reduced in the case of using the greatest common divisor grouping seed filters.

For the $g$, on the other hand, because the parameters and computations depend on the seed filters, different $g$ will not affect the FLOPs and the number of parameters.

\begin{table}
    \centering
    \caption{Ablation Studies of Groups $g$ in GroupNL.}
    \label{ref:gcomp}
    \resizebox{.35\textwidth}{!}{%
    \begin{tabular}{lccccc} \toprule
        \makecell[l]{}& $r$&$g$ & \makecell[c]{\#Params\\(M)} & \makecell[c]{\#FLOPs\\(M)} &\makecell[c]{Top-1\\Acc (\%)}\\ \midrule
        \multirow{6}*{\rotatebox{90}{\makecell[l]{ResNet-18}}}& \multirow{3}*{2} & 4 & 5.6 & 279 & 95.12\\ 
         & & 8 & 5.6 & 279& 94.91\\
         & & 16 & 5.6 & 279& 94.84\\ \cmidrule{2-6}
         & \multirow{3}*{4} & 4 & 2.8 & 141 & 94.33\\ 
         & & 8 & 2.8 & 141 & 94.38 \\
         & & 16 & 2.8 & 141 & 94.15 \\ \midrule
        \multirow{6}*{\rotatebox{90}{\makecell[l]{ResNet-101}}}& \multirow{3}*{2} & 4 & 18.5 & 1159& 95.51\\ 
         & & 8 & 18.5 & 1159& 95.35 \\
         & & 16 & 18.5 & 1159& 95.66 \\ \cmidrule{2-6}
         & \multirow{3}*{4} & 4 & 18.5 & 1159& 95.40\\ 
         & & 8 & 18.5 & 1159& 95.44 \\
         & & 16 & 18.5 & 1159& 95.25 \\\bottomrule 
    \end{tabular}
    }
\end{table}

\subsubsection{Module-Level Profiling}~\label{sec:module_profiling}
We conduct a module-level profiling of GroupNL Conv on Raspberry Pi, based on PyTorch-CPU backend.
For Conv modules, the number of input and output channels is 512, the input feature maps size is (512, 32, 32), kernel size is (3, 3), stride is 2, and padding is 1, so the output feature maps size is (512, 16, 16). 
The FPS reports the average Frames Per Second in 100-times inference with randomly initialized inputs.

\begin{figure}
\centering
\includegraphics[width=1.\linewidth]{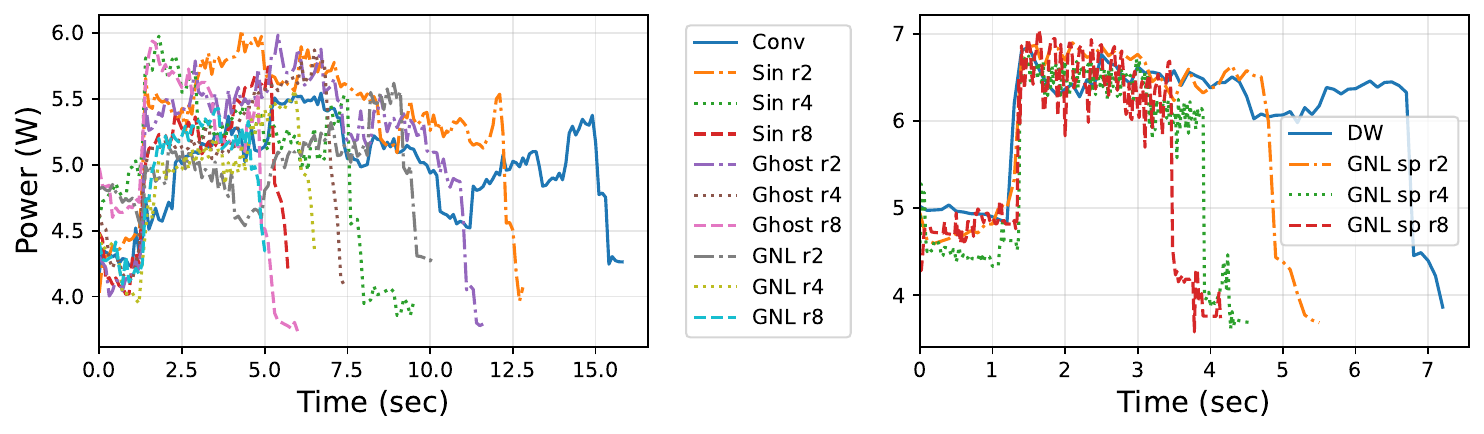}
\caption {Power in Module-level Profiling.}
\label{fig:power_module_profiling}
\end{figure}

\begin{figure}
\centering
\includegraphics[width=1.\linewidth]{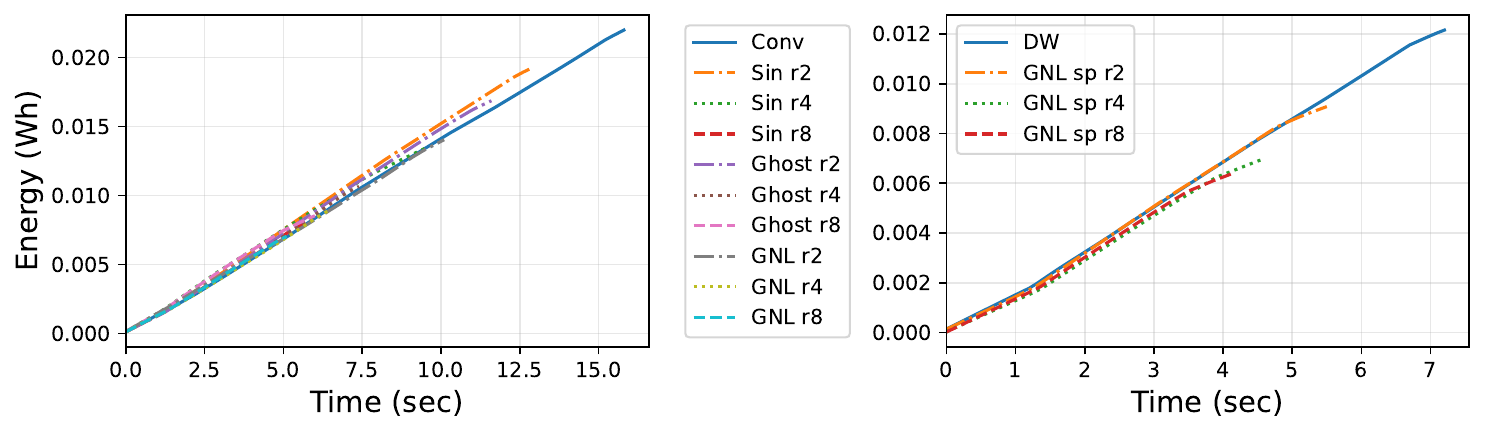}
\caption {Energy Consumption in Module-level Profiling.}
\label{fig:energy_module_profiling}
\end{figure}

As shown by the profiling results in Table~\ref{ref:model_level_pofiling}, GroupNL Conv demonstrates superior efficiency across all ratios $r=2,4,8$. 
Under comparable parameter counts and FLOPs, it consistently achieves the \emph{highest FPS} and the \emph{lowest energy consumption} among all Conv modules.
For example, at $r\!=\!2$, GroupNL Conv is 1.23$\times$ faster than the Ghost Conv while consuming 20.5\% less energy.
Notably, sparse GroupNL Conv achieves performance that is comparable to depthwise Conv, but with fewer parameters (e.g., at $r\!=\!8$, 4.66K vs. 5.12K), significantly higher FPS (53.14 vs. 20.53), and lower energy consumption (0.0056Wh vs. 0.0116Wh).
The power and energy consumption of module-level profiling are reported in Figure~\ref{fig:power_module_profiling} and Figure~\ref{fig:energy_module_profiling}, respectively.
\begin{table}
    \centering
    \caption{Module-level Profiling on Raspberry Pi with PyTorch-CPU Backend.}
    \label{ref:model_level_pofiling}
    \resizebox{.5\textwidth}{!}{%
    \begin{tabular}{cccccccc} \toprule
        \makecell[c]{Module}& 
        $r$ & 
        \makecell[c]{\#Params} & 
        \makecell[c]{\#FLOPs\\(M)} & 
        \makecell[c]{Avg.Times\\(ms)} &
        \makecell[c]{FPS\\($\uparrow$)} & 
        \makecell[c]{Power\\(W)} & 
        \makecell[c]{Energy\\(Wh)} 
        \\ \midrule
        \multirow{1}*{\makecell[c]{Conv}} 
        & - & 2.36M & 603.98 & 122.2 & 8.18 & 5.003 & 0.0208 \\ \midrule
        \multirow{3}*{\makecell[c]{SineFM\\Conv}} 
        & 2 & 1.18M & 302.97 & 97.1 & 10.3 & 5.396 & 0.0183 \\ 
         & 4 & 0.59M & 151.81 & 55.36 & 18.06 & 5.285 & 0.0112\\
         & 8 & 0.30M & 76.23 & 34.86 & 28.69 & 5.024 & 0.0073 \\\midrule
        \multirow{3}*{\makecell[c]{Ghost\\Conv}} 
        & 2 & 1.18M & 302.58 & 86.31 & 11.59 & 5.262 & 0.0161  \\
         & 4  & 0.59M & 151.88 & 49.95 & 20.02 & 5.156 & 0.0099 \\
         & 8 & 0.30M & 76.53 & 31.61 & 31.63 & 5.292 & 0.0072  \\\midrule
        \multirow{3}*{\makecell[c]{GroupNL\\Conv}} 
        & 2 & 1.18M & 302.06 & 70.14 & 14.26 & 5.005 & 0.0128 \\ 
         & 4 & 0.59M & 151.09 & 42.32 & 23.63 & 4.905 & 0.0082 \\
         & 8 & 0.29M & 75.61 & 28.43 & 35.17 & 4.905 & 0.0063 \\ \midrule
         {\makecell[c]{Depthwise Conv}} 
         & -  & 5.12K & 1.18 & 48.71 & 20.53 & 6.120 & 0.0116  \\ \midrule
         \multirow{3}*{\makecell[c]{GroupNL\\Conv (sparse)}} 
         & 2  & 4.62K & 1.25 & 30.68 & 32.6 & 6.096 & 0.0081 \\
         & 4 & 4.63K & 1.28 & 22.65 & 44.16 & 5.730 & 0.0062 \\
         & 8 & 4.66K & 1.29 & 18.82 & 53.14 & 5.831 & 0.0056 \\
         \bottomrule 
    \end{tabular}
    }
\end{table}

\begin{table*}[ht]
\centering
\caption{Severity Levels Comparison in EfficientNet-ES Structure (Top: Vanilla; Down: GroupNL) on ImageNet-C.\label{tab:imagenetc_severitylevel}}
\resizebox{1.\textwidth}{!}{%
\begin{tabular}{@{\hspace{2mm}}ccccccccccccccccccccc@{\hspace{2mm}}}
\toprule
\multirow{3}*{\makecell[c]{Level}}
&\multirow{3}*{\makecell[c]{mCE\\($\downarrow$)}}
&\multicolumn{18}{c}{Corruption Error ($\downarrow$)}\\ \cmidrule{3-21}
&& \multirow{2}*{\makecell[c]{Gauss.\\Noise}} 
& \multirow{2}*{\makecell[c]{Shot\\Noise}} 
& \multirow{2}*{\makecell[c]{Impulse\\Noise}} 
& \multirow{2}*{\makecell[c]{Defocus\\Blur}} 
& \multirow{2}*{\makecell[c]{Glass\\Blur}} 
& \multirow{2}*{\makecell[c]{Motion\\Blur}} 
& \multirow{2}*{\makecell[c]{Zoom\\Blur}} 
& \multirow{2}*{\makecell[c]{Snow}} 
& \multirow{2}*{\makecell[c]{Frost}} 
& \multirow{2}*{\makecell[c]{Fog}} 
& \multirow{2}*{\makecell[c]{Bright.}} 
& \multirow{2}*{\makecell[c]{Contrast}} 
& \multirow{2}*{\makecell[c]{Elastic\\Trans.}} 
& \multirow{2}*{\makecell[c]{Pixelate}} 
& \multirow{2}*{\makecell[c]{JPEG\\Comp.}}
& \multirow{2}*{\makecell[c]{Speckle\\Noise}}
& \multirow{2}*{\makecell[c]{Gauss.\\Blur}}
& \multirow{2}*{\makecell[c]{Spatter}}
& \multirow{2}*{\makecell[c]{Saturate}}
\\
\\ \midrule
\multirow{2}*{\makecell[c]{Mean}} & 55.033 & 62.052 & 63.339 & 65.818 & 64.430 & 73.497 & 58.716 & 65.290 & 59.704 & 54.999 & 47.713 & 30.732 & 45.922 & 57.111 & 60.785 & 44.755 & 54.082 & 60.322 & 41.914 & 34.451
\\
& \textbf{53.958} & \textcolor{blue}{60.843} & \textcolor{blue}{61.825} & \textcolor{blue}{65.581} & \textcolor{blue}{62.466} & \textcolor{blue}{72.968} & \textcolor{blue}{58.113} & \textcolor{blue}{63.870} & \textcolor{blue}{57.496} & \textcolor{blue}{54.063} & \textcolor{blue}{47.302} & \textcolor{blue}{30.174} & 46.991 & \textcolor{blue}{55.622} & \textcolor{blue}{58.412} & \textcolor{blue}{43.345} & \textcolor{blue}{53.372} & \textcolor{blue}{58.514} & \textcolor{blue}{40.574} & \textcolor{blue}{33.670}

\\ \midrule 
\multirow{2}*{\makecell[c]{Severity 1}} & 36.260 & 36.416 & 36.568 & 39.068 & 43.558 & 46.032 & 34.822 & 48.798 & 40.4 & 35.408 & 34.764 & 27.09 & 31.504 & 32.602 & 36.878 & 35.164 & 35.152 & 33.166 & 28.818 & 32.724 
\\
& \textbf{35.599} & \textcolor{blue}{35.996} & \textcolor{blue}{36.492} & 40.402 & \textcolor{blue}{42.060} & \textcolor{blue}{44.626} & \textcolor{blue}{33.856} & \textcolor{blue}{47.136} & \textcolor{blue}{38.78} & \textcolor{blue}{34.846} & \textcolor{blue}{34.686} & \textcolor{blue}{26.522} & \textcolor{blue}{31.214} & \textcolor{blue}{31.780} & \textcolor{blue}{35.662} & \textcolor{blue}{34.420} & 35.50 & \textcolor{blue}{32.576} & \textcolor{blue}{27.910} & \textcolor{blue}{31.910}

\\ \midrule 
\multirow{2}*{\makecell[c]{Severity 2}} & 45.169 & 43.776 & 46.122 & 52.036 & 51.022 & 59.184 & 43.698 & 58.968 & 60.264 & 49.080 & 38.620 & 28.162 & 34.250 & 53.546 & 43.180 & 38.784 & 39.126 & 46.816 & 35.982 & 35.586
\\
& \textbf{44.314} & \textcolor{blue}{43.382} & \textcolor{blue}{45.804} & 53.748 & \textcolor{blue}{49.220} & \textcolor{blue}{58.044} & \textcolor{blue}{42.622} & \textcolor{blue}{57.376} & \textcolor{blue}{57.550} & \textcolor{blue}{48.358} & 39.076 & \textcolor{blue}{27.544} & 34.332 & \textcolor{blue}{51.996} & \textcolor{blue}{41.150} & \textcolor{blue}{37.810} & 39.920 & \textcolor{blue}{45.180} & \textcolor{blue}{34.616} & \textcolor{blue}{34.238}

\\ \midrule 
\multirow{2}*{\makecell[c]{Severity 3}} & 53.486 & 58.540 & 60.116 & 61.688 & 65.410 & 83.766 & 58.614 & 66.520 & 56.594 & 59.804 & 45.486 & 29.682 & 38.798 & 48.584 & 54.960 & 41.356 & 54.754 & 61.130 & 42.196 & 28.240
\\
& \textbf{52.404} & \textcolor{blue}{57.376} & \textcolor{blue}{58.850} & 62.140 & \textcolor{blue}{63.276} & \textcolor{blue}{83.656} & \textcolor{blue}{57.852} & \textcolor{blue}{65.074} & \textcolor{blue}{54.288} & \textcolor{blue}{58.764} & 45.746 & \textcolor{blue}{29.134} & 39.670 & \textcolor{blue}{46.558} & \textcolor{blue}{52.248} & \textcolor{blue}{40.040} & \textcolor{blue}{54.288} & \textcolor{blue}{58.622} & \textcolor{blue}{40.414} & \textcolor{blue}{27.676}

\\ \midrule 
\multirow{2}*{\makecell[c]{Severity 4}} & 64.426  & 77.188 & 81.608 & 81.218 & 76.78 & 87.732 & 74.308 & 73.076 & 66.678 & 62.122 & 51.794 & 32.464 & 51.208 & 63.836 & 77.158 & 49.130 & 64.962 & 73.144 & 45.842 & 33.838
\\
& \textbf{62.845} & \textcolor{blue}{75.034} & \textcolor{blue}{78.738} & \textcolor{blue}{78.738} & \textcolor{blue}{74.538} & \textcolor{blue}{87.688} & \textcolor{blue}{74.090} & \textcolor{blue}{71.694} & \textcolor{blue}{63.424} & \textcolor{blue}{60.934} & \textcolor{blue}{50.998} & \textcolor{blue}{31.958} & 52.784 & \textcolor{blue}{61.382} & \textcolor{blue}{72.984} & \textcolor{blue}{47.538} & \textcolor{blue}{63.078} & \textcolor{blue}{70.662} & \textcolor{blue}{44.416} & \textcolor{blue}{33.374}

\\ \midrule 
\multirow{2}*{\makecell[c]{Severity 5}} & 75.826  & 94.338 & 92.28 & 95.080 & 85.380 & 90.772 & 82.136 & 79.088 & 74.586 & 68.580 & 67.90 & 36.262 & 73.848 & 86.988 & 91.750 & 59.340 & 76.416 & 87.356 & 56.734 & 41.868
\\
& \textbf{74.628}  & \textcolor{blue}{92.428} & \textcolor{blue}{89.24} & \textcolor{blue}{92.878} & \textcolor{blue}{83.236} & 90.826 & 82.146 & \textcolor{blue}{78.068} & \textcolor{blue}{73.436} & \textcolor{blue}{67.412} & \textcolor{blue}{66.002} & \textcolor{blue}{35.710} & 76.956 & \textcolor{blue}{86.392} & \textcolor{blue}{90.014} & \textcolor{blue}{56.918} & \textcolor{blue}{74.072} & \textcolor{blue}{85.530} & \textcolor{blue}{55.514} & \textcolor{blue}{41.152}

\\  \bottomrule
\end{tabular}%
}
\end{table*}

\subsubsection{Analysis of Corruption Levels} 
We report the results of five corruption levels for EfficientNet-ES in Table~\ref{tab:imagenetc_severitylevel}.
With very few exceptions (e.g., contrast), GroupNL EfficientNet-ES demonstrates superior robustness across all five severity levels of ImageNet-C corruptions. 
It achieves a lower mCE and outperforms the vanilla model in the vast majority of corruption types at every severity level. 
This consistent improvement highlights GroupNL's effectiveness in enhancing model robustness against a wide range of real-world image distortions.

\section{Conclusion and Future Work} \label{ref-conclusion}
This paper presented a low-resource and robust CNN design method, \ourmethod{}.
In \ourmethod{}, seed feature maps are grouped and nonlinear transformation functions configured with different hyperparameters and lightweight tensor manipulation operations are used to generate diverse feature maps on demand to reduce the number of parameters and FLOPs.
Experimental results show that \ourmethod{} can effectively improve the robustness of CNNs while acceleration the training and inference.
We will analyze the selection of NLFs from a theoretical perspective in our future work.
Specifically, we plan to analyze the selection of NLFs from a theoretical perspective based function approximation theory, and through more fine-grained experimental analysis, such as analyzing the distance of the feature maps generated by each NLF.

\bibliographystyle{IEEEtran}
\bibliography{main}

\end{document}